\documentclass[sigconf]{acmart}
\AtBeginDocument{%
  }

\usepackage{xspace}
\usepackage{graphicx}
\usepackage{pifont}
\usepackage{multirow, makecell}
\usepackage[table,xcdraw]{xcolor}
\usepackage{subcaption}
\usepackage{bm}
\usepackage{enumitem}
\usepackage[ruled,vlined,linesnumbered]{algorithm2e}
\usepackage{environ} 
\usepackage{pifont}

\usepackage{tcolorbox}
\tcbuselibrary{skins, breakable}

\newtcolorbox[auto counter]{findingbox}{
    breakable,
    enhanced,
    before skip=3pt,          
    after skip=3pt,
    top=2pt,                  
    bottom=2pt,
    left=3pt,                 
    right=3pt,
    pad at break=1pt,         
    bottomsep at break=0pt,   
    colback=gray!10,          
    colframe=black,           
    boxrule=1.1pt,            
    sharp corners,            
    fontupper=\itshape,       
    title={Finding \thetcbcounter.},
    colbacktitle=gray!10,
    coltitle=black,
    fonttitle=\bfseries\upshape,
    attach title to upper,
    after title={\hspace{0.5em}} 
}

\newcommand{\hi}[1]{\vspace{.25em} \noindent {\bf #1} }

\newcommand{\llm}{\textsc{LLM}\xspace}
\newcommand{\llms}{\textsc{LLMs}\xspace}

\newcommand{\blue}[1]{\textcolor{blue}{#1}}

\newcommand{\lgl}[1]{\textcolor{blue}{LGL #1 LGL}}
\newcommand{\zw}[1]{\textcolor{purple}{#1}}
\newcommand{\zxh}[1]{\textcolor{blue}{#1}}

\setcopyright{none}
\renewcommand\footnotetextcopyrightpermission[1]{} 
\begin{document}

\title{Memory as Data: An In-depth Study of Agent Memory Methods: \\ \protect [Experiments \& Analysis]}
\title{Are We Ready For An Agent-Native Memory System? [E\&A]}
\title{Are We Ready For An Agent-Native Memory System?}

\settopmatter{authorsperrow=4}

\author{Wei Zhou}
\affiliation{
    Shanghai Jiao Tong University
    \country{}
}
\email{weizhoudb@sjtu.edu.cn}

\author{Xuanhe Zhou}\authornote{Xuanhe Zhou is the corresponding author.}
\affiliation{
    Shanghai Jiao Tong University
    \country{}
}
\email{zhouxuanhe@sjtu.edu.cn}

\author{Shaokun Han}
\affiliation{
    Shanghai Jiao Tong University
    \country{}
}
\email{areedd0@sjtu.edu.cn}

\author{Hongming Xu}
\affiliation{
    Shanghai Jiao Tong University
    \country{}
}
\email{muzhihai@sjtu.edu.cn}

\author{Guoliang Li}
\affiliation{
    \institution{Tsinghua University}
    \city{}
    \country{}
}
\email{liguoliang@tsinghua.edu.cn}

\author{Zhiyu Li}
\affiliation{
    MemTensor (Shanghai) Technology Co., Ltd
    \country{}
}
\email{lizy@memtensor.cn}

\author{Feiyu Xiong}
\affiliation{
    MemTensor (Shanghai) Technology Co., Ltd
    \country{}
}
\email{xiongfy@memtensor.cn}

\author{Fan Wu}
\affiliation{
    Shanghai Jiao Tong University
    \country{}
}
\email{fwu@cs.sjtu.edu.cn}

\renewcommand{\shortauthors}{Trovato et al.}

\pagestyle{plain}
\pagenumbering{arabic}

\begin{abstract}
Memory for large language model (LLM) agents has rapidly evolved from simple retrieval-augmented mechanisms into a data management system that supports persistent information storage, retrieval, update, consolidation, and dynamic lifecycle governance throughout agent execution. Despite this evolution, existing evaluations still benchmark agent memory mainly through end-to-end task success metrics (e.g., F1, BLEU), while treating the underlying system as a monolithic black box. As a result, critical system-level concerns, including operational costs, architectural trade-offs across memory modules, and robustness under dynamic knowledge updates, remain insufficiently explored.


In this paper, we present a systematic experimental study of agent memory from a data management perspective. We propose an analytical framework that decomposes agent memory into four core modules: memory representation and storage, extraction, retrieval and routing, and maintenance. Under this framework, we evaluate 12 representative memory systems and two reference baselines across five benchmark workloads spanning 11 datasets. Our extensive end-to-end evaluation shows that no single architecture dominates across all scenarios; instead, effectiveness depends heavily on how well the memory structure aligns with the workload bottleneck. Furthermore, through fine-grained ablation studies, we quantify their individual effects on representation fidelity, retrieval precision, update correctness, and long-horizon stability. Finally, we reveal cost-performance trade-offs under realistic workloads, showing localized maintenance is more cost-efficient than global reorganization. Based on these findings, we identify promising directions towards building truly agent-native memory systems.
The code is publicly available at \blue{\emph{\url{https://github.com/OpenDataBox/MemoryData}}}.


\end{abstract}

\maketitle

\section{Introduction}
\label{sec:intro}

The rapid evolution of Large Language Model (LLM) agents has sparked a large body of exciting research and industrial efforts in building agent memory, i.e., the data management system of the LLM agent that supports long-horizon stateful execution and personalized interaction~\cite{luo2026data,khan2026rag,liu2026supporting,singh2024personal,openai2026agents,microsoft2025copilot,google2025adk}. 

\begin{figure}[!t]
  \centering
  \includegraphics[width=\linewidth]{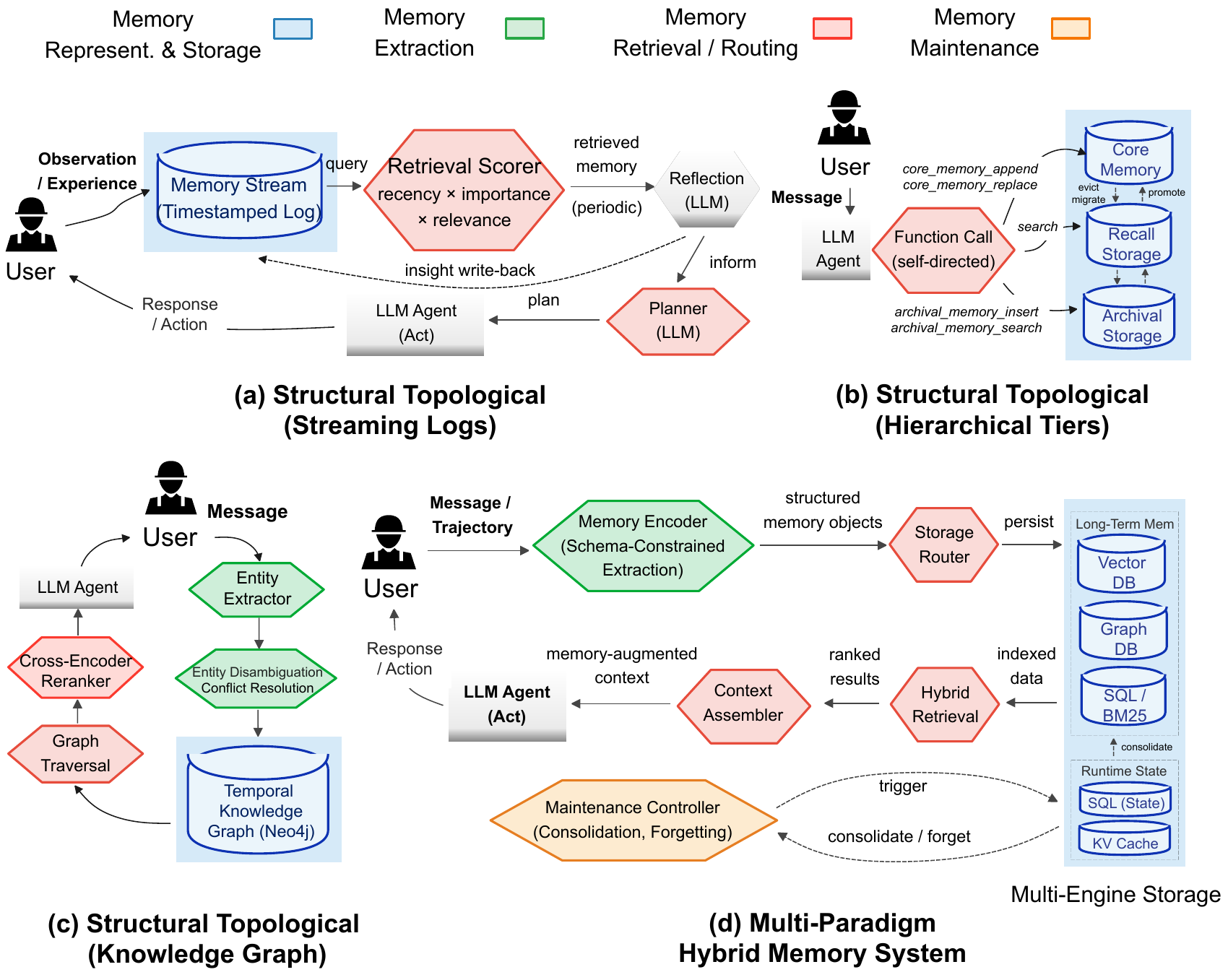}
  \vspace{-.5cm}
  \caption{Typical Execution Workflows of Agent Memory.}
  \label{fig:intro}
  \vspace{-.5cm}
\end{figure}

As shown in Figure~\ref{fig:intro}, existing agent memory systems span a diverse set of architectural designs. 
\textsf{(1) Stream-and-Reflection Memory System} (e.g., MemoryBank~\cite{memorybank}) maintains experiences as timestamped memory streams and periodically summarizes them into higher-level reflections that are written back into the stream; 
\textsf{(2) Hierarchical Tiered Memory System} (e.g., MemGPT~\cite{packer2023memgpt}) organizes memory into multiple levels with different capacities and access properties, separating core memory from archival storage with explicit movement (e.g., eviction and promotion) across tiers; 
\textsf{(3) Knowledge Graph Memory System} (e.g., $\text{Mem0}^g$~\cite{chhikara2025mem0}, Zep~\cite{rasmussen2025zep}) represents entities, relations, and their temporal evolution in structured forms (e.g., temporal knowledge graphs), often incorporating entity disambiguation and conflict resolution; \textsf{(4) Composite Hybrid Memory System} (e.g., A-MEM~\cite{xu2025amem}) routes schema-aware memory objects across multiple storage substrates, explicitly separating runtime state (e.g., KV caches) from long-term storage (e.g., vector, graph, keyword indexes), managed by dedicated maintenance modules. However, this rapid proliferation has also led to a highly fragmented landscape that lacks systematic evaluation from a data management perspective, raising a natural question: \textit{Are we ready for an agent-native memory system?}


In this paper, we revisit this question for agent memory. In particular, we focus on \textit{system-level memory} over textual, structured, and even parametric representations~\cite{chhikara2025mem0,packer2023memgpt,rasmussen2025zep,xu2025amem}, a fundamental infrastructure component for modern autonomous agents.
We focus on memory-centric systems, rather than task-specific agent frameworks where memory is an auxiliary module~\cite{claudecode, zhou2026dbcooker}.
It is the persistent data management system that maintains information beyond a single inference step (e.g., historical interactions, environmental observations, and intermediate tool executions) decoupled from the LLMs' parametric weights and volatile context windows. Agent frameworks rely on these external memory systems (e.g., Mem0~\cite{chhikara2025mem0}, Letta~\cite{packer2023memgpt}, Zep~\cite{rasmussen2025zep}, and A-MEM~\cite{xu2025amem}) to actively write, update, index, and route relevant context back into the reasoning loop. The capability of a long-horizon agent largely depends on the reliability and efficiency of this memory layer. An agent adopting a poorly designed memory architecture can suffer from factual contradictions, catastrophic forgetting, or unacceptable latencies during continuous execution~\cite{du2026memory,zheng2025lifelong}.

Recent benchmarks~\cite{maharana2024locomo,wu2024longmemeval,memoryagentbench2026,membench} have evaluated agent memory and shown that external memory can improve agent performance on tasks requiring factual recall and long-context understanding. However, these evaluations are largely rooted in natural language processing and have multiple limitations when treating agent memory as a data management system (see Section~\ref{sec:def}). 
First, they fail to evaluate many representative memory architectures (e.g., systems such as MemoChat, MemTree, and LightMem have not been included in prior evaluations) under unified workloads, making principled cross-system comparisons difficult. Efforts from the database community~\cite{wu2026memoryeab} limit their scope to a few chatbot-centric datasets (e.g., LoCoMo and LongMemEval only), neglecting complex agentic execution scenarios. 
Second, existing benchmarks predominantly rely on single-sided, end-to-end task success metrics (e.g., F1 and BLEU scores) rather than a comprehensive evaluation suite. They fail to explicitly isolate and measure multi-dimensional performance indicators such as evidence-level retrieval fidelity, dynamic update robustness under conflicting knowledge, and long-horizon stability. 
Third, they rarely measure key operational costs from a systems perspective, such as index construction time and query latency, which are critical for production deployments.
Last, they treat memory systems as monolithic black boxes rather than decomposing them into fundamental data management modules for isolated, fine-grained analysis.

We overcome these limitations and conduct comprehensive experiments and analyses from a data management perspective. The contributions are as follows.

\noindent
\textbf{(1) Technology Decomposition and Taxonomy (Section~\ref{sec:overview}).}
We decompose existing agent memory systems into four core components: (i) memory representation and storage, (ii) memory extraction, (iii) memory retrieval and routing, and (iv) memory maintenance. For each component, we further establish a structured taxonomy\footnote{\blue{\url{https://github.com/OpenDataBox/awesome-agent-memory}}} by categorizing existing approaches according to their underlying design principles, enabling principled comparisons.

\noindent
\textbf{(2) Overall End-to-End Performance Evaluation (Section~\ref{sec:overall}).}
We conduct end-to-end evaluations under a unified and fair testbed (e.g., {unified time-overhead traces})\footnote{\blue{\url{https://github.com/OpenDataBox/MemoryData}}} across five distinct benchmark workloads encompassing 11 datasets. Our study includes 12 representative memory systems, each embodying different combinations of representation, storage, routing, and maintenance strategies. We evaluate their performance from five perspectives: task effectiveness (RQ1), retrieval fidelity (RQ2), dynamic update robustness (RQ3), long-horizon stability (RQ4), and operational cost (RQ5).

\noindent
\textbf{(3) Fine-Grained Technical Component Evaluation (Section~\ref{sec:finegrained}).}
Leveraging our four-module framework, we conduct controlled and fine-grained experiments on representative strategies within each technique component. By systematically generating controlled variants that modify one module at a time, we quantify their performance trade-offs and assess their individual impacts on representation fidelity, routing precision, and update correctness.

\noindent
\textbf{(4) Insightful Findings.}
Based on our experimental results and in-depth analysis, we distill a set of insightful findings regarding the cost--performance trade-offs of agent memory systems:

\noindent
\ding{182} \textbf{Are Memory Systems Effective Across Different Agent Request Workloads?} No single memory architecture dominates all scenarios. Composite hybrid systems lead on conversational QA, while graph-based methods excel in single-hop factual recall but struggle with temporal reasoning. Moreover, effective memory systems remain robust across \llm backbone variants because they externalize evidence localization before answer generation.

\noindent
\ding{183} \textbf{How Accurately Do Memory Systems Retrieve Stored Evidence?} Explicit query planning and balanced hybrid search maximize contextual relevance. However, retrieval accuracy degrades significantly as the temporal distance between the evidence and the query increases, exposing limitations of similarity-based retrieval.

\noindent
\ding{184} \textbf{Are Memory Systems Robust Under Dynamic Updates?} Graph-based methods handle knowledge updates most reliably, whereas popular fact-extraction plugins and append-only stores struggle with targeted overwrites. Systems lacking lifecycle management return stale facts, leading to ``hallucinations of the past''.

\noindent
\ding{185} \textbf{Do Memory Systems Remain Stable Over Long Horizons?} Many append-only memory stores suffer from catastrophic degradation as evidence becomes more distant. For time-dependent queries, raw long-context retrieval still outperforms most memory-backed approaches, indicating that standard semantic consolidation often destroys crucial chronological cues.

\noindent
\ding{186} \textbf{What Are the Operational Costs of Agent Memory?} Highly structured systems incur orders-of-magnitude higher index construction time and query latency than lightweight stores, yet do not consistently deliver proportional accuracy gains.

\noindent
\ding{187} \textbf{When Do Individual Memory Components Go Wrong?} Each layer of abstraction (e.g., compression, summarization, and fact extraction) progressively discards information. Furthermore, fine-grained \llm-based extraction can yield modest precision gains but substantially degrade multi-hop reasoning. Finally, conservative memory consolidation serves as the best default maintenance strategy, whereas delayed flushing creates a deceptive trade-off between surface-level coverage and actual answerability.

\section{Preliminaries}
\label{sec:def}

To support the discussion in the rest of this paper, we first clarify the scope of \emph{agent memory} from a data management perspective. Although recent studies have examined memory from viewpoints such as cognitive taxonomy, agent architecture, and graph-based organization~\cite{zhang2024survey,du2026memory,hu2025memory,wu2026memoryeab,tang2026survey,yang2026graph}, the underlying concept is still often treated primarily as an algorithmic component of the LLM or agent pipeline~\cite{zhang2024survey,du2026memory,hu2025memory}. In contrast, we study agent memory as a standalone data management object and system infrastructure, with explicit attention to how it is represented, stored, retrieved, updated, and maintained under real agent workloads. Under this view, we introduce a set of definitions below.


\begin{table*}[!t]
\vspace{-1.25cm}
\caption{Taxonomy and Characteristics of Agent Memory Systems.}
\vspace{-.35cm}
\label{tab:taxonomy}
\resizebox{\linewidth}{!}{
\begin{tabular}{|c|l|ll|l|l|l|}
\hline
\rowcolor[HTML]{1F1F1F} 
\cellcolor[HTML]{1F1F1F}{\color[HTML]{FFFFFF} }                                             & \multicolumn{1}{c|}{\cellcolor[HTML]{1F1F1F}{\color[HTML]{FFFFFF} }}                                  & \multicolumn{2}{c|}{\cellcolor[HTML]{1F1F1F}{\color[HTML]{FFFFFF} \textbf{Memory Representation \& Storage}}}                                                                                                                                                                           & \multicolumn{1}{c|}{\cellcolor[HTML]{1F1F1F}{\color[HTML]{FFFFFF} }}                                                            & \multicolumn{1}{c|}{\cellcolor[HTML]{1F1F1F}{\color[HTML]{FFFFFF} }}                                                                                                       & \multicolumn{1}{c|}{\cellcolor[HTML]{1F1F1F}{\color[HTML]{FFFFFF} }}                                                               \\ \cline{3-4}
\rowcolor[HTML]{1F1F1F} 
\multirow{-2}{*}{\cellcolor[HTML]{1F1F1F}{\color[HTML]{FFFFFF} \textbf{Category}}}          & \multicolumn{1}{c|}{\multirow{-2}{*}{\cellcolor[HTML]{1F1F1F}{\color[HTML]{FFFFFF} \textbf{Method}}}} & \multicolumn{1}{c|}{\cellcolor[HTML]{1F1F1F}{\color[HTML]{FFFFFF} Representation}}                                                                  & \multicolumn{1}{c|}{\cellcolor[HTML]{1F1F1F}{\color[HTML]{FFFFFF} \textbf{Storage}}}                                              & \multicolumn{1}{c|}{\multirow{-2}{*}{\cellcolor[HTML]{1F1F1F}{\color[HTML]{FFFFFF} \textbf{Memory Extraction}}}}                & \multicolumn{1}{c|}{\multirow{-2}{*}{\cellcolor[HTML]{1F1F1F}{\color[HTML]{FFFFFF} \textbf{\begin{tabular}[c]{@{}c@{}}Memory Retrieval \&\\ Query Routing\end{tabular}}}}} & \multicolumn{1}{c|}{\multirow{-2}{*}{\cellcolor[HTML]{1F1F1F}{\color[HTML]{FFFFFF} \textbf{Memory Maintenance}}}}                  \\ \hline
                                                                                            & \textbf{MemoChat}~\cite{Memochat}                                                                                    & \multicolumn{1}{l|}{\begin{tabular}[c]{@{}l@{}}\ding{182} Token-Level Sequence\\ (Structured JSON Memos)\end{tabular}}             & \ding{182} Transient In-Context Registers                                                                        & \begin{tabular}[c]{@{}l@{}}\ding{184} Schema-Constrained Extraction\\ (LLM Topic Segmentation)\end{tabular}    & \begin{tabular}[c]{@{}l@{}}\ding{185} Autonomous Agentic Routing\\ (LLM Topic Selection)\end{tabular}                                                     & \begin{tabular}[c]{@{}l@{}}\ding{184} LLM-Driven Semantic Consolidation\\ (Turn-Triggered)\end{tabular}           \\ \cline{2-7} 
                                                                                            & \textbf{Mem0}~\cite{chhikara2025mem0}                                                                                         & \multicolumn{1}{l|}{\begin{tabular}[c]{@{}l@{}}\ding{182} Token-Level Sequence\\ (Discrete Facts)\end{tabular}}                    & \begin{tabular}[c]{@{}l@{}}\ding{183} Specialized Single-Engine\\ (Vector DB)\end{tabular}                       & \ding{183} Schema-Free Extraction                                                                              & \ding{183} Semantic-Based Retrieval                                                                                                                       & \begin{tabular}[c]{@{}l@{}}\ding{184} LLM-Driven Semantic Consolidation\\ (Tool-Calling)\end{tabular}             \\ \cline{2-7} 
                                                                                            & \textbf{MEM1}~\cite{mem1}                                                                                         & \multicolumn{1}{l|}{\ding{182} Token-Level Sequence}                                                                               & \ding{182} Transient In-Context Registers                                                                        & \ding{182} Raw Sequence Concatenation                                                                          & \ding{182} Native Attention-Based Retrieval                                                                                                               & \ding{183} Capacity-Driven Physical Eviction                                                                      \\ \cline{2-7} 
\multirow{-7}{*}{\textbf{\begin{tabular}[c]{@{}c@{}}Sequential\\ Context\end{tabular}}}     & \textbf{MemAgent}~\cite{memagent}                                                                                     & \multicolumn{1}{l|}{\ding{182} Token-Level Sequence}                                                                               & \ding{182} Transient In-Context Registers                                                                        & \begin{tabular}[c]{@{}l@{}}\ding{182} Raw Sequence Concatenation\\ (Recursive Summaries)\end{tabular}          & \ding{182} Native Attention-Based Retrieval                                                                                                               & \begin{tabular}[c]{@{}l@{}}\ding{183} Capacity-Driven Physical Eviction\\ (RL Overwrite)\end{tabular}             \\ \hline
                                                                                            & \textbf{MemTree}~\cite{MemTree}                                                                                      & \multicolumn{1}{l|}{\begin{tabular}[c]{@{}l@{}}\ding{183} Graph \& Tree-Based Topology\\ (Hierarchical Tree)\end{tabular}}         & \begin{tabular}[c]{@{}l@{}}\ding{183} Specialized Single-Engine\\ (Vector DB)\end{tabular}                       & \begin{tabular}[c]{@{}l@{}}\ding{183} Schema-Free Extraction\\ (Top-Down Embedding)\end{tabular}               & \begin{tabular}[c]{@{}l@{}}\ding{183} Semantic-Based Retrieval\\ (Collapsed Tree)\end{tabular}                                                            & \begin{tabular}[c]{@{}l@{}}\ding{184} LLM-Driven Semantic Consolidation\\ (Recursive Aggregation)\end{tabular}    \\ \cline{2-7} 
                                                                                            & \textbf{Zep}~\cite{rasmussen2025zep}                                                                                          & \multicolumn{1}{l|}{\begin{tabular}[c]{@{}l@{}}\ding{183} Graph \& Tree-Based Topology\\ (Temporal KG)\end{tabular}}               & \begin{tabular}[c]{@{}l@{}}\ding{183} Specialized Single-Engine\\ (Graph DB)\end{tabular}                        & \begin{tabular}[c]{@{}l@{}}\ding{184} Schema-Constrained Extraction\\ (Triplets)\end{tabular}                  & \begin{tabular}[c]{@{}l@{}}\ding{186} Multi-Stage Hybrid Execution\\ (Dense + BM25 + BFS)\end{tabular}                                                    & \begin{tabular}[c]{@{}l@{}}\ding{182} Timestamp-Based Multi-Versioning\\ (Logical Invalidation)\end{tabular}      \\ \cline{2-7} 
                                                                                            & $\textbf{Mem0}^g$~\cite{chhikara2025mem0}                                                                                        & \multicolumn{1}{l|}{\begin{tabular}[c]{@{}l@{}}\ding{183} Graph \& Tree-Based Topology\\ (Labeled Graph)\end{tabular}}             & \begin{tabular}[c]{@{}l@{}}\ding{184} Heterogeneous Multi-Engine\\ (Vector + Graph DB)\end{tabular}              & \begin{tabular}[c]{@{}l@{}}\ding{184} Schema-Constrained Extraction\\ (Entity-Relation)\end{tabular}           & \ding{184} Topological Subgraph Traversal                                                                                                                 & \ding{182} Timestamp-Based Multi-Versioning                                                                       \\ \cline{2-7} 
\multirow{-8}{*}{\textbf{\begin{tabular}[c]{@{}c@{}}Structural\\ Topological\end{tabular}}} & \textbf{Cognee}~\cite{cognee}                                                                                       & \multicolumn{1}{l|}{\begin{tabular}[c]{@{}l@{}}\ding{183} Graph \& Tree-Based Topology \\ (Entity-Relation Triplets)\end{tabular}} & \begin{tabular}[c]{@{}l@{}}\ding{184} Heterogeneous Multi-Engine\\ (Graph + Vector + Relational DB)\end{tabular} & \begin{tabular}[c]{@{}l@{}}\ding{184} Schema-Constrained Extraction\\ (ECL Pipeline via Pydantic)\end{tabular} & \begin{tabular}[c]{@{}l@{}}\ding{184} Topological Subgraph Traversal\\ (Dense-Seeded Triplet Extraction)\end{tabular}                                     & \begin{tabular}[c]{@{}l@{}}\ding{182} Timestamp-Based Multi-Versioning \\ (Hash-Based Deduplication)\end{tabular} \\ \hline
                                                                                            & \textbf{LightMem}~\cite{LightMem}                                                                                     & \multicolumn{1}{l|}{\begin{tabular}[c]{@{}l@{}}\ding{184} Heterogeneous Composite\\ (Tripartite Schema)\end{tabular}}              & \begin{tabular}[c]{@{}l@{}}\ding{183} Specialized Single-Engine\\ (Relational DB)\end{tabular}                   & \begin{tabular}[c]{@{}l@{}}\ding{183} Schema-Free Extraction\\ (Entropy-Gated)\end{tabular}                    & \ding{183} Semantic-Based Retrieval                                                                                                                       & \begin{tabular}[c]{@{}l@{}}\ding{182} Timestamp-Based Multi-Versioning\\ (Append-Only Logs)\end{tabular}          \\ \cline{2-7} 
                                                                                            & \textbf{SimpleMem}~\cite{simplemem}                                                                                    & \multicolumn{1}{l|}{\ding{184} Heterogeneous Composite}                                                                            & \begin{tabular}[c]{@{}l@{}}\ding{184} Heterogeneous Multi-Engine\\ (Vector DB + BM25 + SQL)\end{tabular}         & \ding{184} Schema-Constrained Extraction                                                                       & \begin{tabular}[c]{@{}l@{}}\ding{185} Autonomous Agentic Routing\\ (Query Expansion)\end{tabular}                                                         & \begin{tabular}[c]{@{}l@{}}\ding{184} LLM-Driven Semantic Consolidation\\ (On-the-Fly Synthesis)\end{tabular}     \\ \cline{2-7} 
                                                                                            & \textbf{MemOS}~\cite{MemOS}                                                                                        & \multicolumn{1}{l|}{\begin{tabular}[c]{@{}l@{}}\ding{184} Heterogeneous Composite\\ (MemCube)\end{tabular}}                        & \begin{tabular}[c]{@{}l@{}}\ding{184} Heterogeneous Multi-Engine\\ (Vector + Graph DB)\end{tabular}              & \begin{tabular}[c]{@{}l@{}}\ding{184} Schema-Constrained Extraction\\ (Semantic Parser)\end{tabular}           & \begin{tabular}[c]{@{}l@{}}\ding{186} Multi-Stage Hybrid Execution\\ (Boolean + Semantic)\end{tabular}                                                    & \begin{tabular}[c]{@{}l@{}}\ding{182} Timestamp-Based Multi-Versioning\\ (Differential Writes)\end{tabular}       \\ \cline{2-7} 
                                                                                            & \textbf{MemoryOS}~\cite{memoryos}                                                                                     & \multicolumn{1}{l|}{\begin{tabular}[c]{@{}l@{}}\ding{184} Heterogeneous Composite\\ (Segment-Page)\end{tabular}}                   & \begin{tabular}[c]{@{}l@{}}\ding{184} Heterogeneous Multi-Engine\\ (Keyword Index + Vector DB)\end{tabular}      & \ding{184} Schema-Constrained Extraction                                                                       & \begin{tabular}[c]{@{}l@{}}\ding{186} Multi-Stage Hybrid Execution\\ (Hierarchical Routing)\end{tabular}                                                  & \begin{tabular}[c]{@{}l@{}}\ding{183} Capacity-Driven Physical Eviction\\ (Heat-Based Eviction)\end{tabular}      \\ \cline{2-7} 
                                                                                            & \textbf{A-MEM}~\cite{xu2025amem}                                                                                        & \multicolumn{1}{l|}{\begin{tabular}[c]{@{}l@{}}\ding{184} Heterogeneous Composite\\ (Atomic Notes)\end{tabular}}                   & \begin{tabular}[c]{@{}l@{}}\ding{184} Heterogeneous Multi-Engine\\ (Vector + Graph DB)\end{tabular}              & \begin{tabular}[c]{@{}l@{}}\ding{184} Schema-Constrained Extraction\\ (JSON Attributes)\end{tabular}           & \ding{184} Topological Subgraph Traversal                                                                                                                 & \begin{tabular}[c]{@{}l@{}}\ding{184} LLM-Driven Semantic Consolidation\\ (Mutation \& Pruning)\end{tabular}      \\ \cline{2-7} 
\multirow{-12}{*}{\textbf{\begin{tabular}[c]{@{}c@{}}Multi-Paradigm\\ Hybrid\end{tabular}}}  & \textbf{Letta}~\cite{packer2023memgpt}                                                                               & \multicolumn{1}{l|}{\begin{tabular}[c]{@{}l@{}}\ding{184} Heterogeneous Composite\\ (Context Tiers)\end{tabular}}                  & \begin{tabular}[c]{@{}l@{}}\ding{183} Specialized Single-Engine\\ (Relational DB)\end{tabular}                   & \ding{184} Schema-Constrained Extraction                                                                       & \begin{tabular}[c]{@{}l@{}}\ding{185} Autonomous Agentic Routing\\ (Function Calling)\end{tabular}                                                        & \begin{tabular}[c]{@{}l@{}}\ding{183} Capacity-Driven Physical Eviction\\ (Queue Flush)\end{tabular}              \\ \hline
\end{tabular}
}
\end{table*}

\hi{Memory Types.} For an LLM agent, a wide variety of information is produced and may need to be memorized, including dialogue history, tool execution logs, distilled facts, and user preferences~\cite{zhang2024survey}. Following established cognitive frameworks, memory can be broadly organized along two axes~\cite{hu2025memory, tang2026survey}. 
(1) Along the \emph{temporal} axis, {short-term memory} holds the volatile state of an ongoing session, while {long-term memory} persists across sessions. 
(2) Along the \emph{functional} axis, long-term memory is further divided into information such as concrete past events (episodic memory), abstracted factual knowledge (semantic memory)~\cite{hu2025memory, wu2026memoryeab}, reusable action strategies (procedural memory), and user preferences.


\hi{Agent Memory.} We define the \emph{agent memory} $\mathcal{M}$ as the persistent data management object~\cite{packer2023memgpt, wu2026memoryeab} that maintains this cumulative state beyond a single inference step and makes it accessible to the agent during future reasoning and action~\cite{zhang2024survey, hu2025memory}.



\hi{Agent Memory System.} To operationalize agent memory $\mathcal{M}$, a robust infrastructure is required~\cite{packer2023memgpt, hu2025memory}. 
As shown in Table~\ref{tab:taxonomy}, from a data systems perspective~\cite{wu2026memoryeab, llm4db}, we formalize the \emph{agent memory system} as a tuple of four modules: $\mathcal{M}_{sys} = \langle \mathcal{R}, \mathcal{S}, \mathcal{Q}, \mathcal{U} \rangle$, where each module governs a distinct phase of the memory lifecycle.

\noindent $\bullet$ \underline{(1) Memory Representation and Storage $\mathcal{R}$:} A mapping that defines the logical and physical memory format with a data model of two facets: (a) logical representation, spanning simple primitives (discrete tokens, continuous vectors) to complex topologies (knowledge graphs, trees, and composites); and (b) physical storage, utilizing transient registers, specialized single-engine databases, or multi-engine backends for persistence and indexing.

\noindent $\bullet$ \underline{(2) Memory Extraction $\mathcal{S}$:} A mechanism governing how heterogeneous input streams (e.g., multi-turn dialogues, tool logs) are transformed into logical memory primitives via pipelines such as raw sequence concatenation, schema-free semantic extraction, or schema-constrained structured extraction.

\noindent $\bullet$ \underline{(3) Memory Retrieval and Routing $\mathcal{Q}$:} A function that dynamically identifies relevant memory subsets based on a query context, utilizing specific routing algorithms to traverse indices. Mechanisms span native attention-based retrieval, semantic $K$-nearest neighbor search, topological subgraph traversal, autonomous agentic routing via \llm planning, and multi-stage hybrid execution.

\noindent $\bullet$ \underline{(4) Memory Maintenance $\mathcal{U}$:} Policies governing the dynamic lifecycle of memory entries, decomposed into three sub-operations: \textit{(a) Conflict Resolution and Versioning} handles contradictions via multi-versioning, invalidation, or precedence rules; \textit{(b) Capacity Management} enforces bounded growth through constraint-based hard eviction (e.g., FIFO, token limits) or score-based priority eviction (e.g., temporal decay); and \textit{(c) Semantic Consolidation} utilizes the \llm to merge redundant assertions into dense summaries or execute CRUD operations via tool-calling interfaces.




\hi{Distinction from RAG and Context Engineering.} 
Retrieval-Augmented Generation (RAG)~\cite{RAG,khan2026rag} typically operates as a stateless, read-only retrieval primitive: given a query, it fetches relevant passages from a static corpus to augment a single generation step. Context engineering~\cite{anthropic2025context} is the broader practice of curating the finite LLM context window at each inference turn (e.g., dynamically selecting prompts, tool descriptions, and retrieved facts) to mitigate context rot~\cite{anthropic2025context}. In contrast, an agent memory system (1) is a persistent and updatable infrastructure for managing agent-specific state over time and (2) governs the full long-term memory lifecycle, including memory representation, storage, retrieval, and maintenance, rather than merely packing the current context window.


\hi{Distinction from Traditional Database Workloads.} Agent memory workloads differ substantially from conventional database OLTP / OLAP workloads~\cite{packer2023memgpt, liu2026supporting}. First, memory access is often \emph{semantic} rather than purely predicate-based~\cite{kang2025bigvectorbench, caminal2025filtered}. 
Queries are commonly expressed through natural language, partial context, or latent intent, and therefore rely on approximate matching, query rewriting, or \llm-guided retrieval rather than only exact logical predicates over rigid schemas. Second, memory contents evolve under \emph{continuous and potentially conflicting observations}. 
Unlike conventional transactional settings, where updates typically overwrite tuples under a predefined schema and consistency model, agent memory must accommodate uncertain, partial, and sometimes contradictory information collected across time, tools, and environments~\cite{hu2025memory, zheng2025lifelong}. Third, agent memory workloads are \emph{highly heterogeneous} in both access pattern and granularity. 
A single workload may combine long-context synthesis, episodic recall, structured fact lookup, temporal reasoning, and streaming updates. 
As a result, practical systems often require hybrid execution strategies that combine semantic retrieval, structured filtering, and topology-aware traversal within one memory architecture~\cite{wu2026memoryeab, packer2023memgpt}. 
These properties distinguish agent memory from traditional databases, motivating dedicated abstractions and evaluation methodologies.

\section{Method Overview}
\label{sec:overview}

In this section, we carefully analyze existing agent memory systems across the four components in Section~\ref{sec:def} and establish a unified taxonomy that summarizes representative component methods.


\subsection{Memory Representation and Storage}
\label{subsec:represent}

\begin{figure}[!t]
  \centering
  \includegraphics[width=\linewidth]{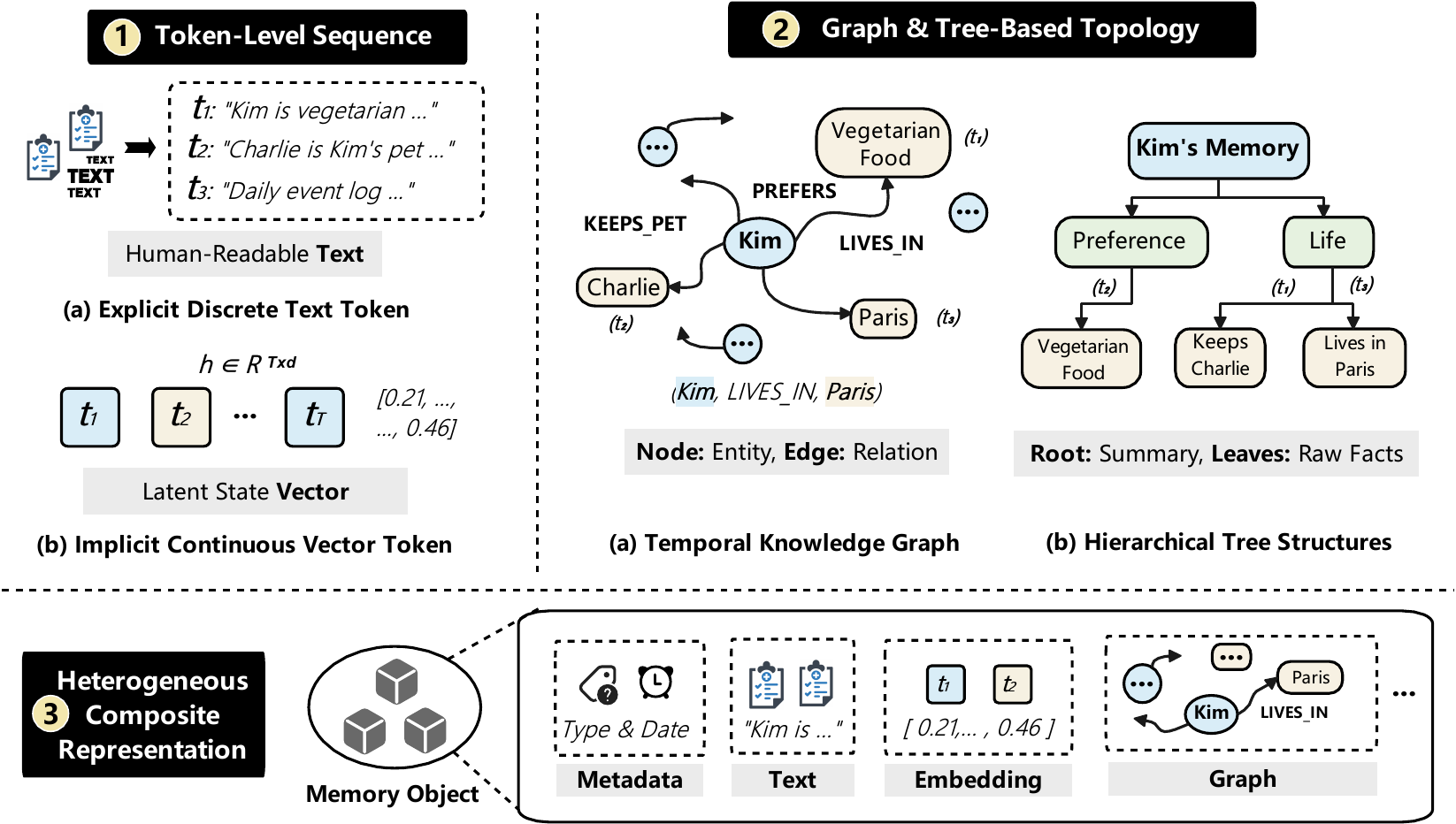}
  \vspace{-.5cm}
  \caption{Memory Representation Methods.}
  \label{fig:memory_representation}
  \vspace{-.5cm}
\end{figure}



This module consists of two components:
(1) logical representation, which defines the structural encoding and organization exposed to the agent system, directly dictating capacity, accessibility, and trade-offs in expressiveness, retrieval granularity, and downstream reasoning compatibility;
and (2) physical storage, which designates the persistence and indexing structures, such as volatile in-context registers, dense vector engines, or topological graph databases.


\subsubsection{\textbf{Logical Representation}}
As displayed in Figure~\ref{fig:memory_representation}, this component acts as a bridge between raw data and the execution environment by organizing memory into clear models, such as graphs, or vector spaces.
It determines how efficiently a system can search, combine, and use historical context for complex tasks.

\noindent \ding{182} \textbf{Token-Level Sequence Representation.} This category models memory as flat, one-dimensional sequences lacking explicit structural abstractions (e.g., graphs or hierarchies). Memory is represented either as discrete, human-readable natural language tokens or as implicit, continuous latent vector tokens (e.g., fact embeddings, hidden states, or KV-cache tensors).

\noindent \underline{\textit{$\blacktriangleright$ Explicit Discrete Text Token.}}
This category models memory as human-readable strings or independent factual statements.
For instance, Mem0 isolates memory into discrete natural language facts extracted directly from interaction history. Similarly, MemoChat structures multi-turn dialogues into discrete JSON blocks (topics, summaries, raw turns) to maintain topical coherence within a plain-text paradigm. While these systems externalize their plain-text memory, others retain it within the active processing window: MemAgent restricts its internal belief state to a strictly bounded text sequence (e.g., 1024 tokens), and MEM1 encapsulates internal-state summaries within specialized boundary tags (e.g., <IS>).

\begin{sloppypar}
\noindent \underline{\textit{$\blacktriangleright$ Implicit Continuous Vector Token.}}
Departing from readable text tokens, this sub-category encodes memory as continuous vectors, which may be materialized either as external embeddings attached to facts and summaries for semantic retrieval, or as model-side latent states such as compressed internal states and attention caches.
For example, Mem0 represents extracted facts as dense semantic embeddings and MemoRAG utilizes specifically initialized weight matrices to compress raw inputs into high-dimensional Key-Value (KV) cache tensors. 
Although these vector-token representations reduce explicit tokenization burdens and integrate naturally with retrieval or inference pipelines, they sacrifice structural interpretability and are difficult to manipulate via fine-grained operations, such as predicate-level filtering or targeted updates to encoded facts.
\end{sloppypar}


\noindent \ding{183} \textbf{Graph and Tree-Based Topological Representation.}
This category abstracts memory into structured graph and tree topologies with interconnected nodes and edges, allowing conversational entities, high-level concepts, and their temporal or semantic relationships to be explicitly modeled and computationally traversed.

\begin{figure}[!t]
  \centering
  \includegraphics[width=\linewidth]{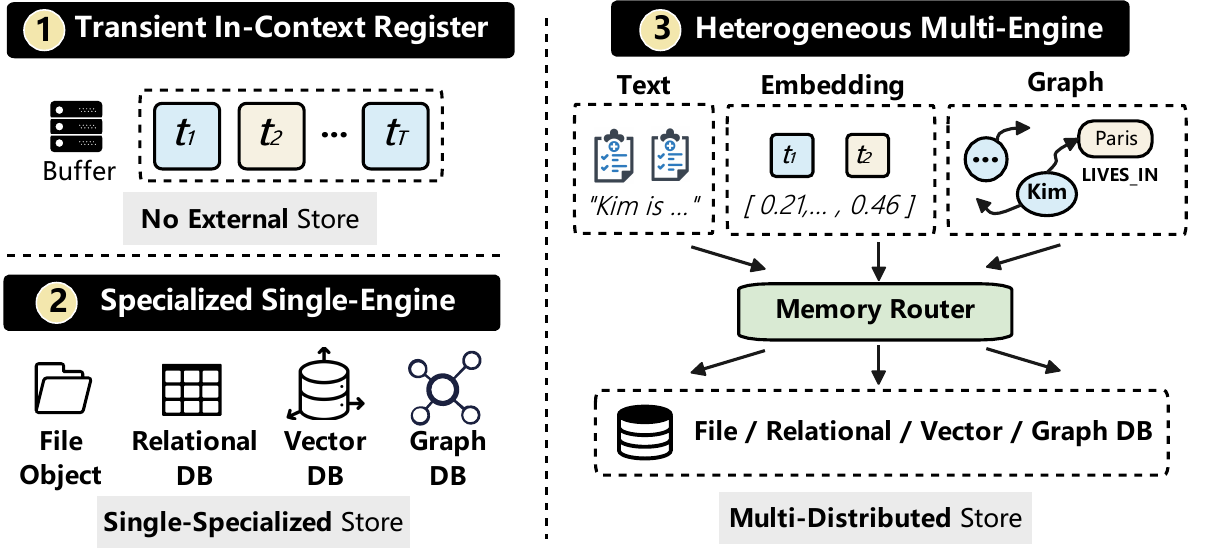}
  \caption{Memory Storage Methods.}
  \label{fig:memory_storage}
  \vspace{-.75cm}
\end{figure}


\noindent \underline{\textit{$\blacktriangleright$ Temporal Knowledge Graphs.}}This sub-category models memory using graph topologies to map entities and their interconnections, natively supporting temporal reasoning and conflict detection.
For example, Zep partitions memory into formally defined, temporally-aware knowledge graphs (e.g., episode, entity, and community subgraphs). Similarly, $\text{Mem0}^g$ formalizes memory as a directed labeled graph, where vertices represent entities and edges encapsulate relationship triplets (e.g., ``LIVES\_IN''). To aid temporal reasoning, entity nodes are enriched with structural metadata like semantic types, dense embeddings, and creation timestamps.

\noindent \underline{\textit{$\blacktriangleright$ Hierarchical Tree Structures.}}
This sub-category organizes knowledge into recursive, hierarchical structures, preserving highly granular observations at terminal leaves and broad semantic abstractions at ancestor nodes. For example, MemTree models memory as a dynamic, directed tree schema. Each node is structured as a tuple containing textual content, a dense embedding, topological pointers, and a depth scalar. Within this topology, deep leaf nodes retain isolated facts (e.g., a player scoring), while ancestor nodes provide high-level conceptual summaries (e.g., the match result), with a specialized root node serving as the definitive entry point.



\noindent \ding{184} \textbf{Heterogeneous Composite Representation.}
This category moves past simple token sequences and standard graphs by packaging memory into complex, multi-part data containers. These architectures directly combine unstructured text with highly structured metadata (e.g., timestamps, categorical labels, vector embeddings, and network links) to form a single functional unit. For example, MemOS proposes the MemCube, a unified data object that organizes memory into three distinct payloads (plain-text, activation, and parametric memory) alongside structured details (e.g., ID tags).




\subsubsection{\textbf{Physical Storage and Indexing}}
As shown in Figure~\ref{fig:memory_storage}, this component manages how data is physically stored and accessed, relying on systems like in-memory caches, files, vector engines, or databases.
It sets the actual capacity limits and determines the speed, throughput, and overall scalability of memory operations.

\noindent \ding{182} \textbf{Transient In-Context Register.}
To eliminate disk I/O and external traversal latency, this category retains memory exclusively within the active hardware state (e.g., dynamic context windows or KV caches).
MemoChat avoids dedicated external memory engines and keeps structured JSON-style memos within the LLM context input during its memorization-retrieval-response loop, while MemAgent directly stores summary tokens as Key-Value (KV) cache tensors via dense positional embeddings.

\begin{figure}[!t]
  \centering
  \includegraphics[width=\linewidth]{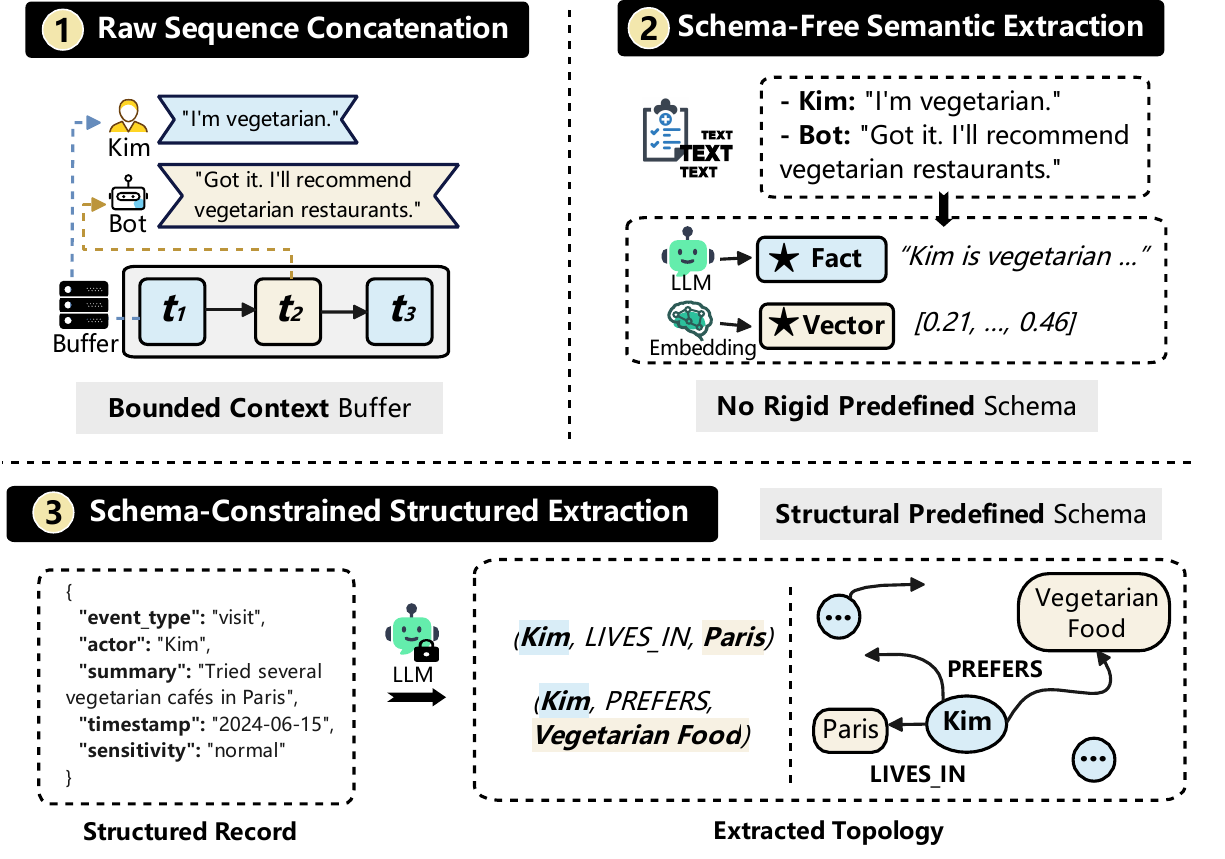}
  \vspace{-.75cm}
  \caption{Memory Extraction Methods.}
  \label{fig:memory_extraction}
  \vspace{-.3cm}
\end{figure}

\noindent \ding{183} \textbf{Specialized Single-Engine Storage.}
This category physically warehouses formulated units within a standalone, homogeneous backend strictly tailored to the memory's logical structure.
Depending on the ingestion paradigm, architectures deploy specific backend topologies: 
(1) \textit{Dense Vector Databases} are utilized to project data into continuous high-dimensional spaces; Mem0 and MemTree use centralized vector stores, Letta leverages PostgreSQL with the \texttt{pgvector} extension;
(2) \textit{Graph Databases} are deployed to enforce topological constraints; both Zep and $\text{Mem0}^g$ execute predefined Cypher queries to physically persist logical graph components into Neo4j;
(3) \textit{Relational SQL Engines} are used to serialize structural and temporal schemas.
LightMem incrementally appends factual streams to preserve global relational states;
(4) \textit{File or Object Stores} preserve raw interaction artifacts (e.g., conversation histories or tool-execution logs) as files or object blobs.


\noindent \ding{184} \textbf{Heterogeneous Multi-Engine Storage.}
This category dynamically constructs multiple index typologies or distributes data across heterogeneous backends (e.g., pairing a dense vector store with a topological graph database). SimpleMem ingests memory into LanceDB with an IVF-PQ mechanism that concurrently maintains dense embeddings, sparse BM25 indices, and SQL predicates. MemoryOS relies on a hybrid index fusing dense cosine similarity with discrete Jaccard similarity. Conversely, MemOS delegates serialized payloads to highly specialized independent backends, fusing Vector and Graph databases via a standardized memory adapter interface.


\subsection{Memory Extraction}
\label{subsec:storage}

Memory extraction concerns how raw interaction traces are computationally processed.
It covers both the extraction pipeline, how language models extract, summarize, or parse unstructured text into logical structures.
As shown in Figure~\ref{fig:memory_extraction}, it defines how the agent memory system transforms heterogeneous input streams (e.g., multi-turn dialogues, and tool execution logs) into logical memory primitives prior to physical persistence.

\noindent \ding{182} \textbf{Raw Sequence Concatenation.}
To minimize computational overhead, this category bypasses explicit extraction prompts, formulating memory directly as raw token concatenations or transient state summaries (e.g., appending recent dialogue turns directly into a prompt buffer).
Systems such as MEM1 and MemAgent retain their newly formulated structures exclusively within the active computational state without secondary parsing.

\noindent \ding{183} \textbf{Schema-Free Semantic Extraction.}
This category systematically distills raw, unstructured inputs into independent, high-value informational units, representing them either as explicit free-form texts or as compressed, continuous latent vectors.
By isolating core knowledge from broader conversational context, it ensures precise and granular retrieval.
For example, Mem0  actively parses interactions to extract and store discrete, standalone factual statements (e.g., ``User is vegetarian and dairy-free'').

\begin{figure}[!t]
  \centering
  \includegraphics[width=\linewidth]{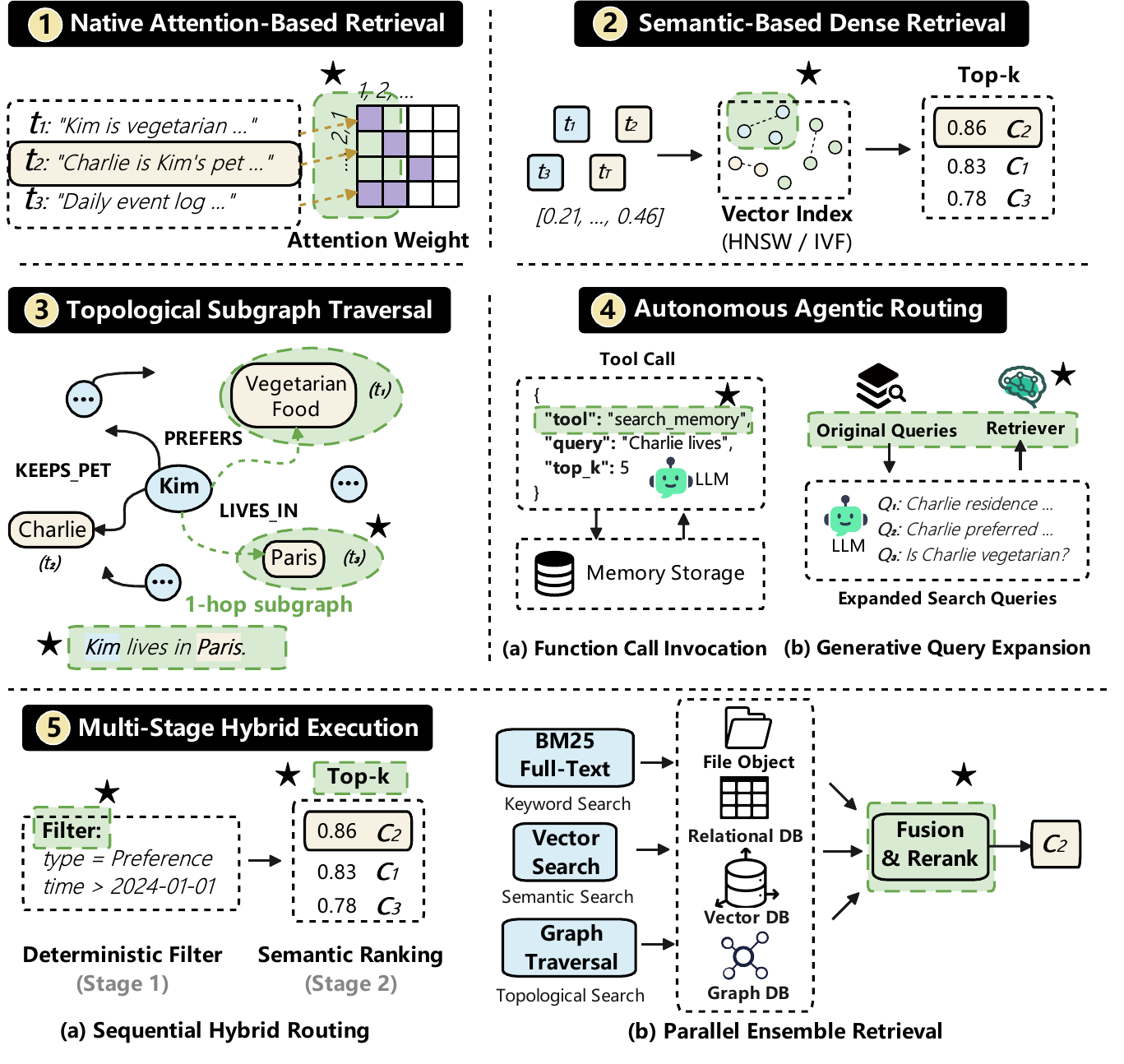}
  \vspace{-.75cm}
  \caption{Memory Retrieval Methods.}
  \label{fig:memory_retrieval}
  \vspace{-.65cm}
\end{figure}

\noindent \ding{184} \textbf{Schema-Constrained Structured Extraction.}
This category prompts the LLM to parse raw inputs and synchronously populate a rigidly predefined structural schema, producing strictly typed data rather than free-form text. The constrained output takes the form of either topological entity-relation triplets for graph insertion or multi-modal relational payloads for hybrid storage, depending on the target backend.
Zep and $\text{Mem0}^g$ extract typed directed relational edges (e.g., \texttt{LIVES\_IN}, \texttt{WORKS\_AT}) conforming to predefined graph schemas, with Zep additionally applying a reflection-inspired verification step to suppress hallucinated triplets.
MemoChat populates predefined structural fields to ensure data predictability by leveraging \llms to segment conversations into strict JSON schemas.



\subsection{Memory Retrieval and Query Routing}
\label{subsec:retrieval}

Memory retrieval and query routing determine how the agent memory system dynamically identifies and extracts relevant historical context to inform the overarching agent's current reasoning state.
As shown in Figure~\ref{fig:memory_retrieval}, this module encompasses the complete query execution spectrum, defining the operational algorithms, predicate evaluations, and agentic workflows utilized to traverse indices. 


\noindent \ding{182} \textbf{Native Attention-Based Retrieval.}
To bypass external database I/O, this category uses the transformer's native computational graph as the sole retrieval engine, relying entirely on self-attention mechanisms to implicitly weight and route information (e.g., scanning dialogue tokens directly within the KV cache). MEM1 performs implicit retrieval via self-attention over the current sequence, utilizing a two-dimensional attention mask to preserve causal consistency. MemAgent implements routing by concatenating blocks directly into the prompt template, enabling standard attention-based decoding without external cross-encoder reranking.

\noindent \ding{183} \textbf{Semantic-Based Dense Retrieval.}
Operating over continuous latent spaces, this category maps query tensors against uniform vector indices to extract localized spatial neighbors (e.g., executing a standard $K$-Nearest Neighbors (KNN) search). Mem0 calculates vector embeddings for incoming queries to execute a dense similarity search, fetching a constrained subset of facts. LightMem utilizes efficient cosine-similarity distance calculations over dense embeddings, bypassing computationally expensive iterative reranking. MemTree implements a collapsed-tree architecture that mathematically flattens its hierarchy, broadcasting inbound vectors to compute global cosine-similarity distributions across all candidates.


\noindent \ding{184} \textbf{Topological Subgraph Traversal.}
Departing from continuous vector spaces, this category retrieves information by traversing explicit relationship edges to extract semantic clusters structurally grounded in knowledge graphs (e.g., hopping from a \texttt{User} node to a linked \texttt{Preference} node).
$\text{Mem0}^g$ deploys an entity-centric heuristic to recursively traverse local subgraphs synchronously with semantic triplet evaluations.
A-MEM identifies candidate anchors via dense $K$-Nearest Neighbor selection, then executes localized graph traversal to access topologically adjacent memory nodes explicitly linked within the same conceptual cluster.

\noindent \ding{185} \textbf{Autonomous Agentic Routing.}
Rather than executing deterministic database scans, this category delegates retrieval to the \llm itself, thereby functioning as an active, autonomous query planner. It generates tool-call invocations or drafts implicit search criteria.

\begin{sloppypar}
\noindent \underline{\textit{$\blacktriangleright$ \textbf{Function Call Invocation}.}}
This sub-category bridges the \llm with external storage by generating explicit function call commands to directly execute predefined database operations (e.g., outputting a valid JSON payload to trigger an external database API). For example, Letta orchestrates self-directed memory retrieval where the LLM evaluates its active context to explicitly generate localized function calls (e.g., emitting an \texttt{archival\_storage.search()} command to extract targeted historical logs).
\end{sloppypar}

\begin{sloppypar}
\noindent \underline{\textit{$\blacktriangleright$ \textbf{Generative Query Expansion}.}}
Unlike rigid function calling, this approach uses natural language generation to synthesize intermediate clues or decompose complex intents before mapping them to the index (e.g., rewriting vague prompts into descriptive search strings). SimpleMem uses an Intent-Aware Retrieval Planning module where the \llm dissects queries, calculates adaptive search depths, and synthesizes optimized query variants.
\end{sloppypar}


\begin{figure}[!t]
  \centering
  \includegraphics[width=\linewidth]{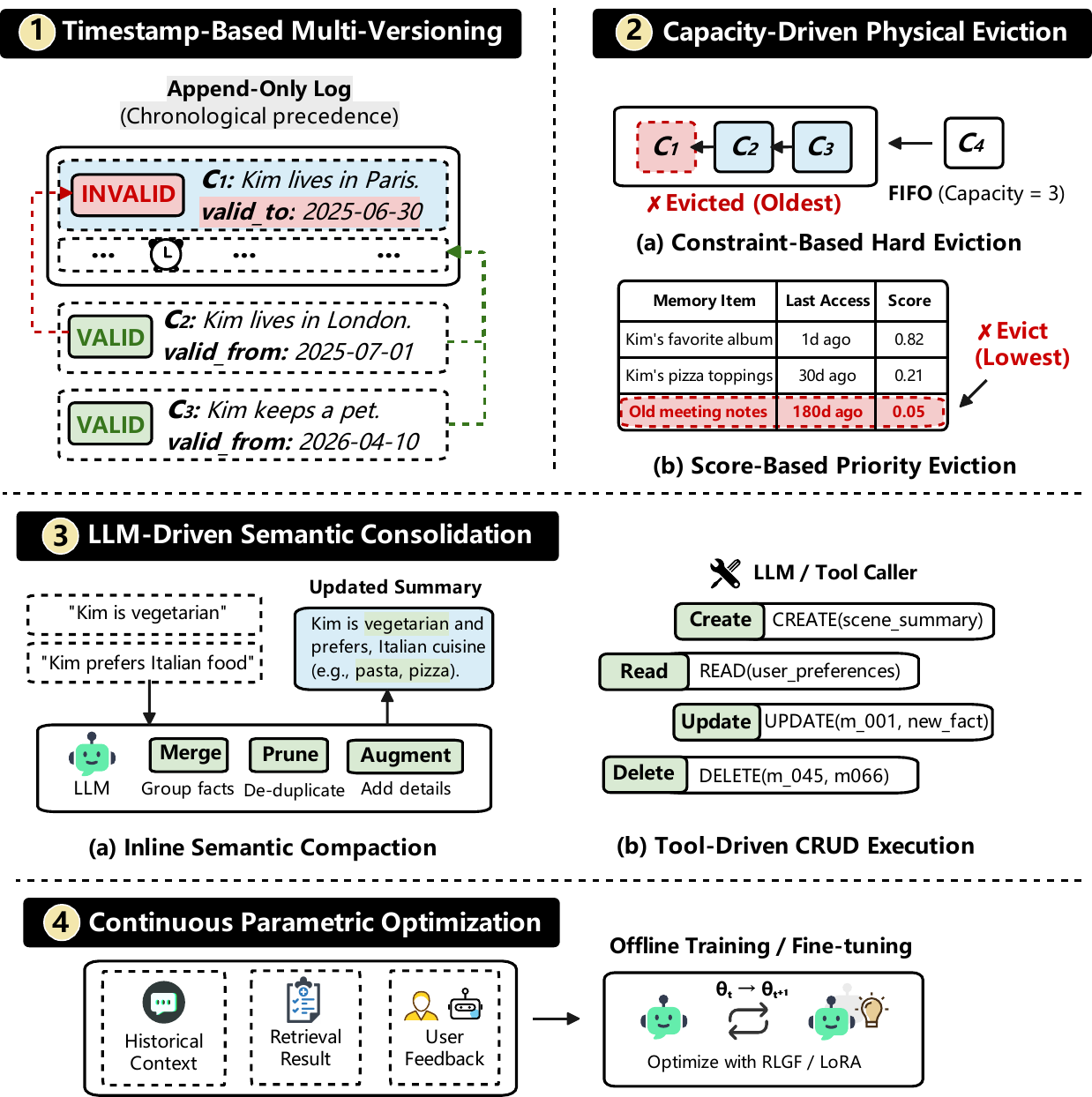}
  \vspace{-.5cm}
  \caption{Memory Maintenance Methods.}
  \label{fig:memory_maintenance}
  \vspace{-.65cm}
\end{figure}

\noindent \ding{186} \textbf{Multi-Stage Hybrid Execution.}
To overcome the recall limitations of single-paradigm searches, this category executes multi-engine query pipelines orchestrating multi-dimensional candidate generation followed by downstream reranking frameworks.

\noindent \underline{\textit{$\blacktriangleright$ \textbf{Sequential Hybrid Routing}.}}
This sub-category chains retrieval paradigms into a strictly ordered pipeline, systematically pruning the search space with deterministic predicates before executing fine-grained semantic extraction (e.g., applying strict SQL date filters before computationally expensive vector searches). MemoryOS executes a federated routing strategy featuring coarse-grained predicate evaluation followed by fine-grained semantic ranking strictly within isolated segments.
It algebraically fuses rule-based structural Boolean filtering with dense semantic similarity routing.

\noindent \underline{\textit{$\blacktriangleright$ \textbf{Parallel Ensemble Retrieval}.}}
In contrast to sequential filtering, this approach maximizes initial recall by simultaneously dispatching queries to multiple distinct indexing algorithms, followed by a late-stage fusion and reranking phase to optimize the aggregated pool (e.g., concurrently fetching candidates via BM25 and dense vector search, then cross-encoding the results). Zep executes simultaneous cosine semantic scans, Okapi BM25 full-text searches, and topological BFS, subsequently optimizing precision via RRF, MMR, and computationally intensive cross-encoder models. 



\subsection{Memory Maintenance}
\label{subsec:maintenance}

Memory maintenance concerns how memory is updated, maintained, compressed, forgotten, and eventually removed over time.
As shown in Figure~\ref{fig:memory_maintenance}, it captures the dynamic behavior of memory after it has been created, including how new information is incorporated, how outdated or conflicting content is revised, and how the system controls memory growth under limited resources.

\begin{figure*}[!t]
\vspace{-.5cm}
\centering
\vspace{-.75cm}
\includegraphics[width=\linewidth]{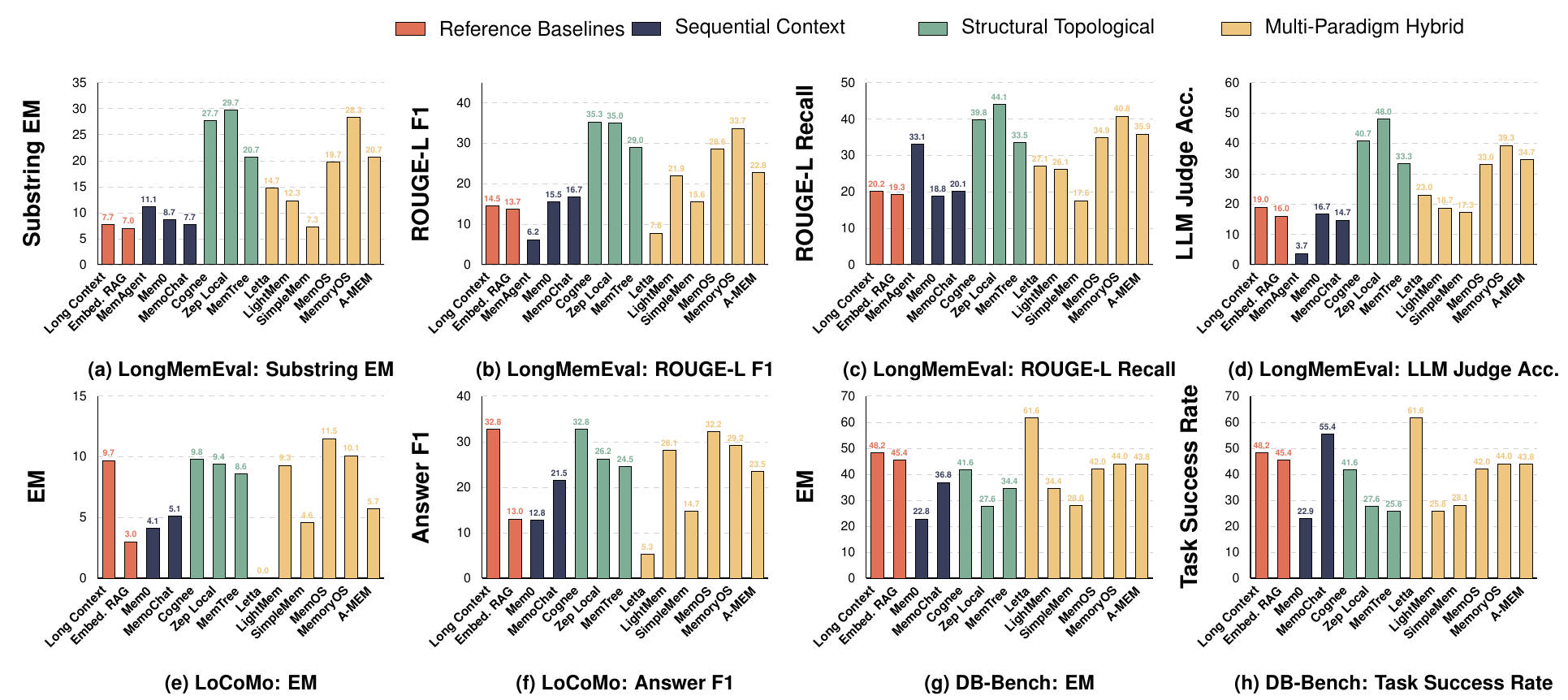}
\caption{Effectiveness of Memory Systems over \textit{LoCoMo}, \textit{MemoryAgentBench (LongMemEval)}, \textit{LifeLongAgentBench (DB-Bench)}.}
\vspace{-.5cm}
\label{fig:RQ1_effectiveness}
\end{figure*}

\noindent \ding{182} \textbf{Timestamp-Based Multi-Versioning.}
Rather than executing physical row deletions, this category preserves historical continuity by utilizing timestamp metadata and append-only logs to logically deprecate expired facts.
Operating via explicit metadata mutations, Zep and $\text{Mem0}^g$ avoid physical deletion by marking obsolete or conflicting relationships as logically invalid using validity flags and timestamps.
Taking an append-only approach, LightMem incrementally inserts timestamped factual streams, while SimpleMem resolves contradictions through strict chronological precedence using ISO-8601 timestamps.
Synthesizing these techniques, MemOS leverages a structured Update API to execute differential writes, seamlessly updating provenance IDs to generate multi-version chains.

\noindent \ding{183} \textbf{Capacity-Driven Physical Eviction.}
In contrast to timestamp-based multi-versioning, this category manages unbounded memory growth by physically dropping or unconditionally overwriting data.
It executes this physical pruning through either strict deterministic constraints or dynamically calculated eviction scores.

\noindent \underline{\textit{$\blacktriangleright$ \textbf{Constraint-Based Hard Eviction}.}}
This sub-category enforces rigid execution bounds by utilizing deterministic rules—such as strict FIFO queues, fixed sequence boundaries, or hard token limits—to unconditionally evict older states.
Executing structural overwrites, MemAgent implements a programmatic scheduling algorithm that unconditionally replaces older memory sequences with newly synthesized summary blocks at every fixed segment boundary. Enforcing hard capacity limits, MEM1 operates through a system-enforced truncation mechanism that executes an automated FIFO pruning protocol to evict older tags once active context thresholds are breached. Operating via threshold flushes, Letta strictly handles buffer capacities via an OS-inspired queue manager; when the token count breaches a terminal limit, it forces a flush sequence to evict older messages into secondary recall storage.

\noindent \underline{\textit{$\blacktriangleright$ \textbf{Score-Based Priority Eviction}.}}
Rather than relying on static capacity limits, this sub-category dynamically forces the physical obsolescence of data by continuously calculating temporal decay or access-frequency scores.
Quantifying access frequency, MemoryOS measures segment vitality via a scalar Heat score that balances retrieval frequency against exponential temporal decay, executing priority evictions that physically target the lowest-heat segments.



\noindent \ding{184} \textbf{\llm-Driven Semantic Consolidation.}
Operating as a cognitive governor, this category leverages the \llm to dynamically resolve logical conflicts and abstract redundant observations into dense summaries prior to query or persistence phases.

\begin{sloppypar}
\noindent \underline{\textit{$\blacktriangleright$ \textbf{Inline Semantic Compaction}.}}
During the active write phase, this sub-category dynamically evaluates and consolidates newly ingested data against existing memory nodes, systematically merging redundant assertions prior to database transaction commitment (e.g., compressing three similar dialogue turns into one dense summary node).
SimpleMem executes online semantic synthesis on-the-fly, systematically merging structurally similar assertions into singular dense abstractions prior to database transaction commitment. MemTree utilizes a core scheduling operation that recursively triggers a semantic summarization prompt across all parent nodes to dynamically fuse historical states with novel payloads.
\end{sloppypar}


\begin{sloppypar}
\noindent \underline{\textit{$\blacktriangleright$ \textbf{Tool-Driven CRUD Execution}.}}
In contrast to automated fusion, this sub-category operationalizes maintenance through discrete, programmed state-mutations guided explicitly by \llm-driven tool interfaces that issue explicit Create, Read, Update, or Delete (CRUD) commands. Mem0 operationalizes its dynamic maintenance strictly through structured \llm tool-calling interfaces encompassing discrete programmed state-mutations such as UPDATE, and DELETE.
\end{sloppypar}


\noindent \ding{185} \textbf{Continuous Parametric Optimization.}
Completely decoupling state updates from online inference latency, this category executes heavy neural optimizations as asynchronous background processes, modifying the actual model parameters rather than the external database schema (e.g., running continuous fine-tuning on overnight batches).
For example, MemoRAG leaves active inference tokens strictly static and read-only, optimizing extraction quality exclusively during an offline training phase via a Reinforcement Learning with Generation Feedback (RLGF) algorithmic framework. 


\section{End-to-End Assessment}
\label{sec:overall}

In this section, we conduct a systematic evaluation of agent memory systems across five research questions.
Across five distinct benchmark workloads and 11 datasets, we assess 12 representative memory systems against baselines to characterize their performance.
Specifically, the five research questions are as follows.


\subsection{Overall Effectiveness (RQ1)}
\label{subsec:effectiveness}

\noindent \textbf{Experimental Setting.}
For \textit{``Do different agent memory systems successfully improve end-to-end task performance across workloads?''}, we evaluate 12 representative memory systems and two reference baselines (\textit{Long Context} and \textit{Embedding RAG}) on the three end-to-end workloads to assess whether memory improves task success beyond the underlying \llm.
Specifically, we use: (1) \textit{LoCoMo}~\cite{maharana2024locomo}: a long-conversation QA benchmark that tests episodic, temporal, and open-domain memory over multi-turn interactions, and report the unweighted mean of category-level \textit{Exact Match (EM)} and \textit{Answer F1} on the four-category queries;
(2) \textit{LongMemEval}~\cite{wu2024longmemeval}: a multi-session long-memory benchmark that evaluates whether systems can reconnect facts across sessions and reason over temporally distributed evidence, and report \textit{Substring EM}, \textit{ROUGE-L F1}, \textit{ROUGE-L Recall}, and \textit{GPT-5.4-based \llm Judge Accuracy} from \textit{MemoryAgentBench}~\cite{memoryagentbench2026}; and (3) \textit{DB-Bench}: evaluates whether memory supports procedural execution across database operations from \textit{LifelongAgentBench}~\cite{DBLP:journals/corr/abs-2505-11942}, and report \textit{Exact Match (EM)} and \textit{Task Success Rate}.

\noindent \textbf{O1-(Cross-Workload Effectiveness): No single memory system dominates all workloads, but methods that preserve task-critical evidence through structure-guided filtering remain the most competitive overall.}
As shown in Figure~\ref{fig:RQ1_effectiveness}, the leading systems shift across workloads: (1) Structure-aware systems lead \textit{LongMemEval}, where Zep reaches \textit{48.0} \textit{\llm Judge Accuracy} and Cognee attains \textit{35.3} \textit{ROUGE-L F1}; 
(2) Hybrid filtering is strongest on \textit{LoCoMo} exactness, where MemOS reaches \textit{11.5} \textit{Exact Match (EM)}; and (3) Trace-preserving memories remain strongest on \textit{DB-Bench}, where Long Context achieves \textit{48.20} \textit{EM} and MemoChat reaches \textit{55.40} \textit{Task Success Rate}.
However, among methods with full workload coverage, MemoryOS and MemOS remain closest to the frontier overall, suggesting that robustness comes not from a single universal memory form, but from preserving the right evidence at the right level of abstraction before final matching.
In particular, (1) Temporal or graph-organized memory is most useful for cross-session aggregation and event-order reasoning (e.g., scattered personal facts in \textit{LongMemEval}); (2) Summary-first or coarse-to-fine routing is useful for exact grounding in long but semantically coherent dialogues (e.g., recovering a specific date or personal detail in \textit{LoCoMo}); and (3) Trace-preserving memory is necessary when correctness depends on intermediate state changes and operation order (e.g., dependent \texttt{UPDATE} and \texttt{INSERT} operations in \textit{DB-Bench}).

\noindent \textbf{O2-(Beyond Exact Match): \textit{EM} remains informative for tasks with canonical, directly grounded outputs, but it becomes insufficient when correctness depends on paraphrastic synthesis or executable success.}
As shown in Figure~\ref{fig:RQ1_effectiveness}, \textit{Exact Match (EM)} is still a meaningful signal on \textit{LoCoMo}, where many questions target short grounded facts, as reflected by MemOS achieving the best \textit{Exact Match (EM)}. 
On \textit{LongMemEval}, however, the stronger systems are more clearly separated once semantic equivalence is considered through \textit{ROUGE-L} and \textit{\llm Judge Accuracy}, indicating that cross-session reasoning often yields correct answers that do not share a single canonical surface form.
On \textit{DB-Bench}, the limitation is even clearer: Long Context achieves the best \textit{Exact Match (EM)}, but MemoChat attains a substantially higher \textit{Task Success Rate}, showing that exact output matching does not fully capture whether memory supports successful execution.
These results suggest that \textit{Exact Match (EM)} is most appropriate when answers are short, canonical, and locally verifiable (e.g., a venue name, or object attribute in \textit{LoCoMo}), but should be complemented once tasks require cross-session synthesis, or end-task state validation (e.g., composing a semantically correct answer from multiple sessions in \textit{LongMemEval} or reaching the correct table state in \textit{DB-Bench}).

\begin{findingbox}
\textbf{(Workload-Aligned Memory).}
RQ1 suggests that strong agent memory is not defined by a single universal representation, but by how well it supports the dominant workload bottleneck: (1) for dispersed cross-session reasoning, relation- and time-aware retrieval is most effective, as in Zep and Cognee; (2) for long but semantically coherent dialogue, coarse-to-fine filtering improves exact grounding, as in MemOS and MemoryOS; and (3) for stateful execution, preserving interaction traces is more critical than exact lexical matching alone, as in Long Context.
\end{findingbox}



\subsection{Memory Retrieval Fidelity (RQ2)}
\label{subsec:retrieval}

\begin{figure}[!t]
  \centering
  \includegraphics[width=\linewidth]{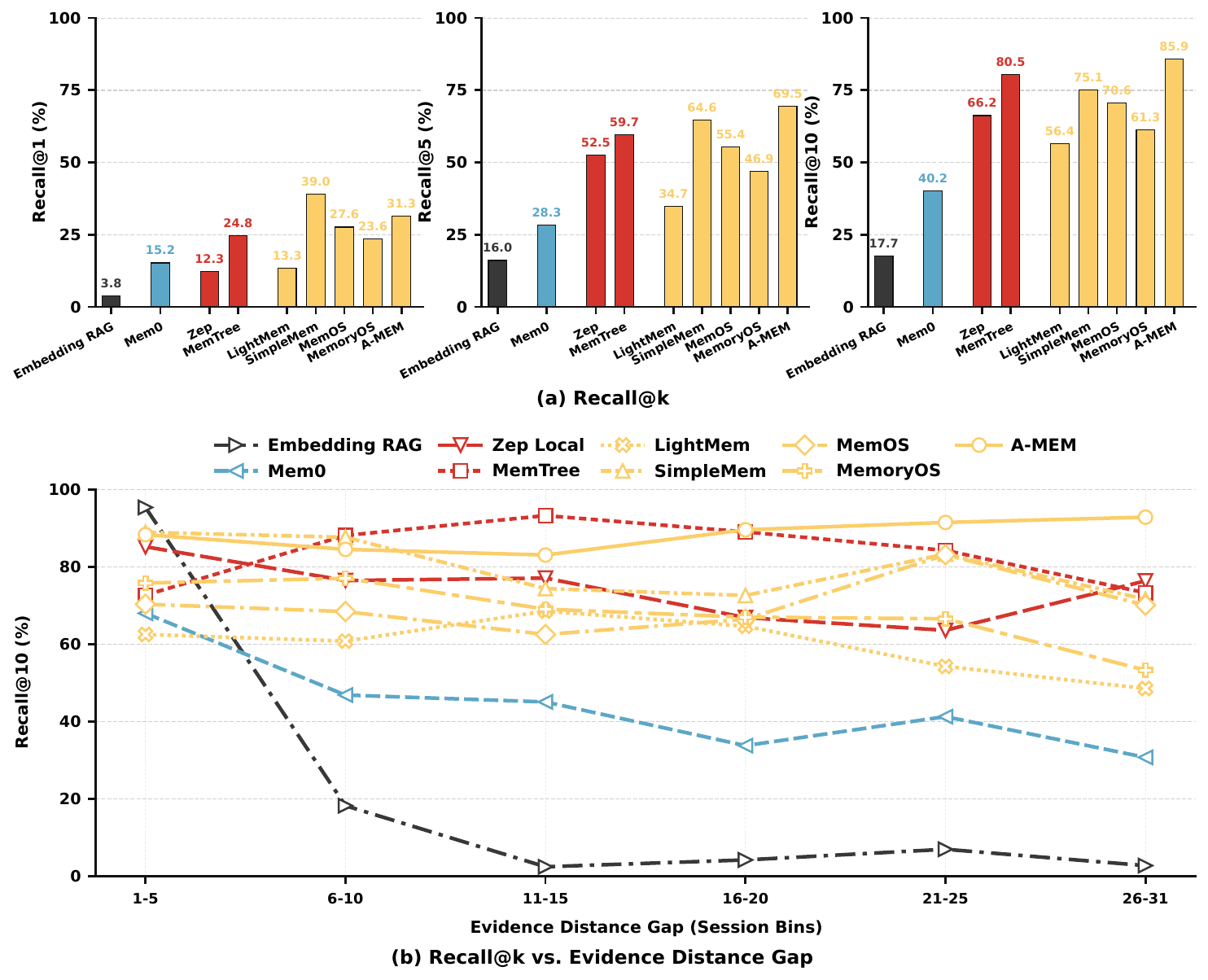}
  \vspace{-.75cm}
  \caption{Retrieval Results of Memory Systems over LoCoMo.}
  \label{fig:RQ2_recall}
  \vspace{-.5cm}
\end{figure}

\noindent \textbf{Experimental Setting.}
For \textit{``How accurately can a memory system surface the stored evidence required by a query?''}, we evaluate eight representative memory systems to assess evidence-level retrieval fidelity independently of downstream answer generation.
Specifically, we use \textit{LoCoMo}~\cite{maharana2024locomo}, which provides source-level gold evidence for queries with diverse evidence distances.
We report: (1) \textit{Recall@K}, where a hit requires the top-$k$ retrieved source-id groups to contain the annotated gold evidence, and (2) \textit{Recall@10} over six \textit{evidence distance gap} bins (1--5 to 26--31), defined by the session distance between the query's final session and the earliest supporting evidence, to measure long-range retrieval accuracy.

\noindent \textbf{O1-(Structured Evidence Expansion): Retrieval fidelity depends less on surfacing one relevant memory early than on preserving an explicitly organized memory structure that can gather complete and temporally distant evidence.}
As shown in Figure~\ref{fig:RQ2_recall}, the results show a clear difference between early-hit precision and overall evidence completeness: \textit{SimpleMem} achieves the highest \textit{Recall@1} (39.0), but \textit{A-MEM} and \textit{MemTree} become clearly stronger at larger retrieval budgets, reaching 69.5/85.9 and 59.7/80.5 on \textit{Recall@5}/\textit{@10}, respectively, while also remaining much more stable as the \textit{evidence distance gap} increases; by contrast, the flat \textit{Embedding RAG} baseline drops sharply after the shortest-gap bin.
This pattern suggests that strong memory retrieval is not mainly a top-1 ranking problem, but an evidence-completion problem in which the required support may be old, scattered, or spread across multiple turns (e.g., personal details mentioned in different sessions or dated events referenced much later).
More specifically, the results point to three different retrieval behaviors:
(1) compression-oriented memory is effective for surfacing one highly relevant item early (e.g., a single salient personal detail or recent conversational fact);
(2) linked or hierarchical memory organization is more effective for gathering complementary supporting evidence across the ranked results (e.g., combining pet names or dated events mentioned in different sessions);
and (3) flat dense retrieval remains competitive mainly when the needed evidence is still close to the current context (e.g., recent conversational facts).
For queries that require scattered or temporally distant support, the most competitive systems are therefore those that organize memory as a structured evidence space rather than a flat similarity cache.

\begin{findingbox}
\textbf{(Evidence-Centric Memory Organization).} RQ2 shows that retrieval quality depends more on how a system organizes evidence for later reconstruction than on how well it ranks one relevant memory first.
Specifically, (1) early localization and evidence assembly should be treated as separate design targets;
(2) explicit structure, such as links or hierarchy, is most valuable when supporting evidence is scattered or temporally distant, as in \textit{A-MEM} and \textit{MemTree};
and (3) flat similarity search is mainly effective for short-range access. 
\end{findingbox}

\subsection{Memory Evolution Robustness (RQ3)}
\label{subsec:robustness}

\begin{table}[t]
\vspace{-1.25cm}
\caption{Robustness over Memory Update Settings.}
\vspace{-.35cm}
\label{tab:RQ3_robustness}
\centering
\footnotesize
\setlength{\tabcolsep}{3.2pt}
\resizebox{\linewidth}{!}{%
\begin{tabular}{ccccccc}
\toprule
\multirow{3}{*}{\textbf{Method}} & \multicolumn{2}{c}{\textbf{LoCoMo}} & \multicolumn{4}{c}{\textbf{LongMemEval}} \\
\cmidrule(lr){2-3} \cmidrule(lr){4-7}
 & \multicolumn{2}{c}{\textbf{Temporal}} & \multicolumn{2}{c}{\textbf{Knowledge Update}} & \multicolumn{2}{c}{\textbf{Temporal Reasoning}} \\
\cmidrule(lr){2-3} \cmidrule(lr){4-5} \cmidrule(lr){6-7}
 & \shortstack{\textbf{Exact} \\ \textbf{Match}} & \shortstack{\textbf{Answer} \\ \textbf{F1}} & \shortstack{\textbf{Substring} \\ \textbf{EM}} & \shortstack{\textbf{ROUGE-L} \\ \textbf{F1}} & \shortstack{\textbf{Substring} \\ \textbf{EM}} & \shortstack{\textbf{ROUGE-L} \\ \textbf{F1}} \\
\midrule
\textbf{Long Context} & 8.1 & 26.9 & 20.0 & 18.0 & 12.0 & 24.0 \\
\textbf{Embedding RAG} & 1.6 & 7.9 & 20.0 & 17.8 & 10.7 & 22.7 \\
\midrule
\textbf{Mem0} & 3.2 & 6.0 & 15.6 & 17.1 & 10.7 & 22.4 \\
\textbf{MemoChat} & 2.4 & 15.4 & 8.9 & 12.9 & 10.7 & 25.3 \\
\midrule
\textbf{Cognee} & 4.0 & \textbf{28.1} & 37.8 & 34.0 & \textbf{18.7} & \textbf{35.8} \\
\textbf{Zep} & 4.8 & 18.1 & \textbf{44.4} & \textbf{36.8} & 13.3 & 30.5 \\
\textbf{MemTree} & 5.6 & 18.6 & 31.1 & 30.6 & 8.0 & 29.9 \\
\midrule
\textbf{Letta (MemGPT)} & 0.0 & 7.1 & 17.8 & 5.7 & 12.0 & 8.8 \\
\textbf{LightMem} & 4.0 & 20.1 & 15.6 & 20.2 & 12.0 & 28.6 \\
\textbf{SimpleMem} & 4.4 & 8.1 & 6.7 & 7.4 & 8.0 & 22.6 \\
\textbf{MemOS} & \textbf{8.9} & 28.0 & 28.9 & 30.5 & 12.0 & 31.1 \\
\textbf{MemoryOS} & 3.2 & 22.7 & 35.6 & 32.2 & 16.0 & 31.6 \\
\textbf{A-MEM} & 4.8 & 17.7 & 26.7 & 22.8 & 8.0 & 22.5 \\
\bottomrule
\end{tabular}
}
\vspace{-.75cm}
\end{table}

\begin{figure}[!t]
    \vspace{-1.25cm}
  \centering
  \includegraphics[width=\linewidth]{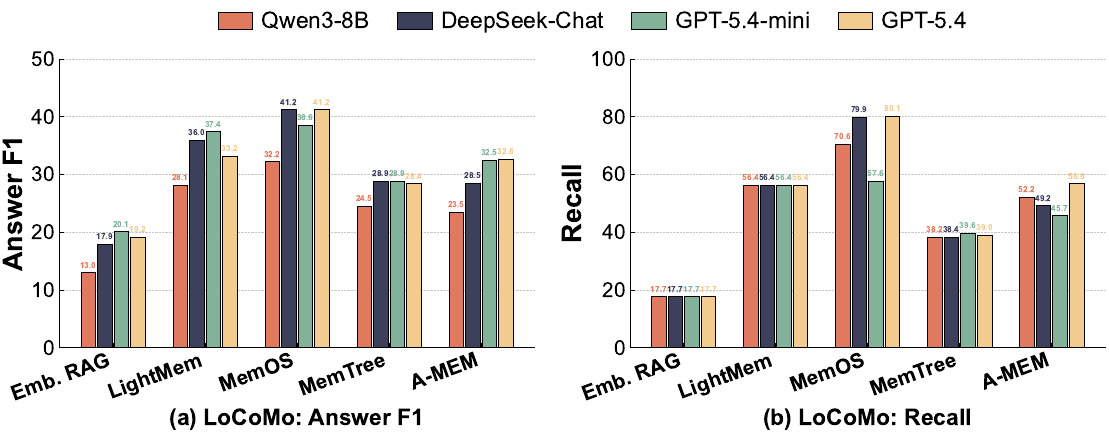}
  \vspace{-.5cm}
  \caption{Ablation of \llm Backbones.}
  \label{fig:RQ3_llm_backbone}
  \vspace{-.83cm}
\end{figure}

\noindent \textbf{Experimental Setting.}
For \textit{``Can agent memory systems reliably incorporate revised facts, preserve the correct temporal state after updates, and remain robust across answer backbones?''}, we conduct two experiments: \textit{(1) Update Robustness Comparison}, which evaluates whether systems can absorb fact revisions and answer temporally grounded queries after updates; and \textit{(2) Backbone Robustness Ablation}, which tests whether this behavior remains stable when only the \llm backbone changes.
In \textit{(1) Update Robustness Comparison}, Table~\ref{tab:RQ3_robustness} compares 11 representative memory systems on \textit{Knowledge Update} and \textit{Temporal Reasoning} from \textit{LongMemEval}~\cite{wu2024longmemeval}, and \textit{Temporal} from \textit{LoCoMo}~\cite{maharana2024locomo}.
The two \textit{LongMemEval}~\cite{wu2024longmemeval} slices are reported with \textit{Substring EM} and \textit{ROUGE-L F1}, while the \textit{LoCoMo}~\cite{maharana2024locomo} slice is reported with \textit{Exact Match (EM)} and \textit{Answer F1};
In \textit{(2) Backbone Robustness Ablation}, Figure~\ref{fig:RQ3_llm_backbone} evaluates 6 representative memory settings under 4 \llm backbones on the \textit{LoCoMo}~\cite{maharana2024locomo}.

\noindent \textbf{O4-(Temporal State Externalization): No single memory system dominates all update-oriented slices, but methods that preserve temporally valid evidence through structured organization remain the most competitive overall.}
As shown in Table~\ref{tab:RQ3_robustness}, the leading systems shift across slices:
(1) Graph- or relation-organized memory is strongest on direct fact revision, where Zep leads \textit{Knowledge Update} with 44.4 \textit{Substring EM} and 36.8 \textit{ROUGE-L F1};
(2) Relationally organized retrieval is strongest on temporally dispersed evidence, where Cognee leads \textit{Temporal Reasoning} with 18.7 \textit{Substring EM} and 35.8 \textit{ROUGE-L F1};
and (3) Hybrid filtered memory is strongest on exact latest-state grounding, where MemOS attains the highest \textit{LoCoMo} \textit{Exact Match (EM)} at 8.9 while Cognee attains the highest \textit{Answer F1} at 28.1.
However, among methods with full slice coverage, Cognee, MemOS, and MemoryOS remain closest to the frontier overall, indicating that robustness comes not from a single universal memory form, but from preserving the right temporal evidence at the right structure level.
In particular, (1) temporal or graph-organized memory is most useful for revised personal facts and dated events (e.g., aggregating scattered updates to preferences, purchases, or past activities in \textit{LongMemEval}); (2) hybrid or coarse-to-fine filtering is most useful when correctness depends on the currently valid state (e.g., recovering the latest date, attribute, or event order from long but semantically coherent dialogues in \textit{LoCoMo}); and (3) flat context accumulation or dense similarity alone is weakest when stale mentions must be separated from updated ones (e.g., distinguishing an earlier personal detail from its later correction after repeated mentions over time).

\noindent \textbf{O5-(Backbone Robustness): Backbone variation changes absolute answer quality more than it changes which memory pipeline remains effective, indicating that stable update behavior is determined primarily before final generation.}
As shown in Figure~\ref{fig:RQ3_llm_backbone}, \textit{Answer F1} generally rises under stronger generators, yet the overall ordering changes only modestly:
(1) MemOS remains the strongest memory-based configuration at 32.2, 41.2, 38.6, and 41.2;
and (2) the only notable reversal is local, where A-MEM overtakes MemTree under GPT-5.4-mini and GPT-5.4.
This stability implies that stronger backbones mostly improve answer realization after relevant evidence has already been localized, rather than compensating for weak temporal grounding.
For example, for date-grounded latest-state queries in the current \textit{LoCoMo} temporal outputs, MemOS remains correct across all four backbones, whereas Embedding RAG remains incorrect.
More concretely, methods with stronger external organization yield a more stable evidence set for different \llms, whereas methods that rely more heavily on \llm-side synthesis exhibit greater cross-backbone movement. 

\begin{findingbox}
\textbf{(Temporal Update Fidelity).}
RQ3 suggests that reliable post-update behavior is a pipeline-level design problem rather than a pure model-capacity problem.
In particular, (1) \emph{revisability} should be built into the memory representation so later facts can be bound to the same entity or event rather than appended as undifferentiated text, as in Zep and Cognee;
(2) \emph{query-time selectivity} should match the workload bottleneck, using filtered or hybrid routing when the task requires the currently valid state, as in MemOS and MemoryOS;
and (3) \emph{\llm scaling} is most valuable only after grounding has succeeded, so stronger backbones should refine answer expression rather than serve as the primary mechanism for resolving stale or conflicting memories.
\end{findingbox}


\subsection{Long Horizon Memory Stability (RQ4)}
\label{subsec:horizon}

\noindent \textbf{Experimental Setting.}
For \textit{``How stable are agent memory systems as the effective memory horizon increases, either through longer contexts or more distant supporting evidence?''}, we evaluate 12 representative memory systems across 3 benchmarks to assess robustness to increasing context length and temporal distance.
Specifically, we use: (1) \textit{LongBench}~\cite{DBLP:conf/acl/BaiLZL0HDLZHDTL24}, which evaluates controlled long-context difficulty in question answering, reported with \textit{Accuracy} over \textit{Short}, \textit{Medium}, and \textit{Long} context-length buckets to measure context-length robustness;
(2) \textit{LongMemEval}~\cite{wu2024longmemeval}, which evaluates multi-session memory as the amount of prior interaction grows, reported with \textit{ROUGE-L F1} over bins of historical session count to measure multi-session stability; and (3) \textit{LoCoMo}~\cite{maharana2024locomo}, which evaluates memory drift when supporting evidence lies back in the conversation, reported with \textit{Answer F1} over bins of evidence-distance gap between the final session and the earliest supporting-evidence.

\begin{figure}[!t]
  \centering
  \includegraphics[width=\linewidth]{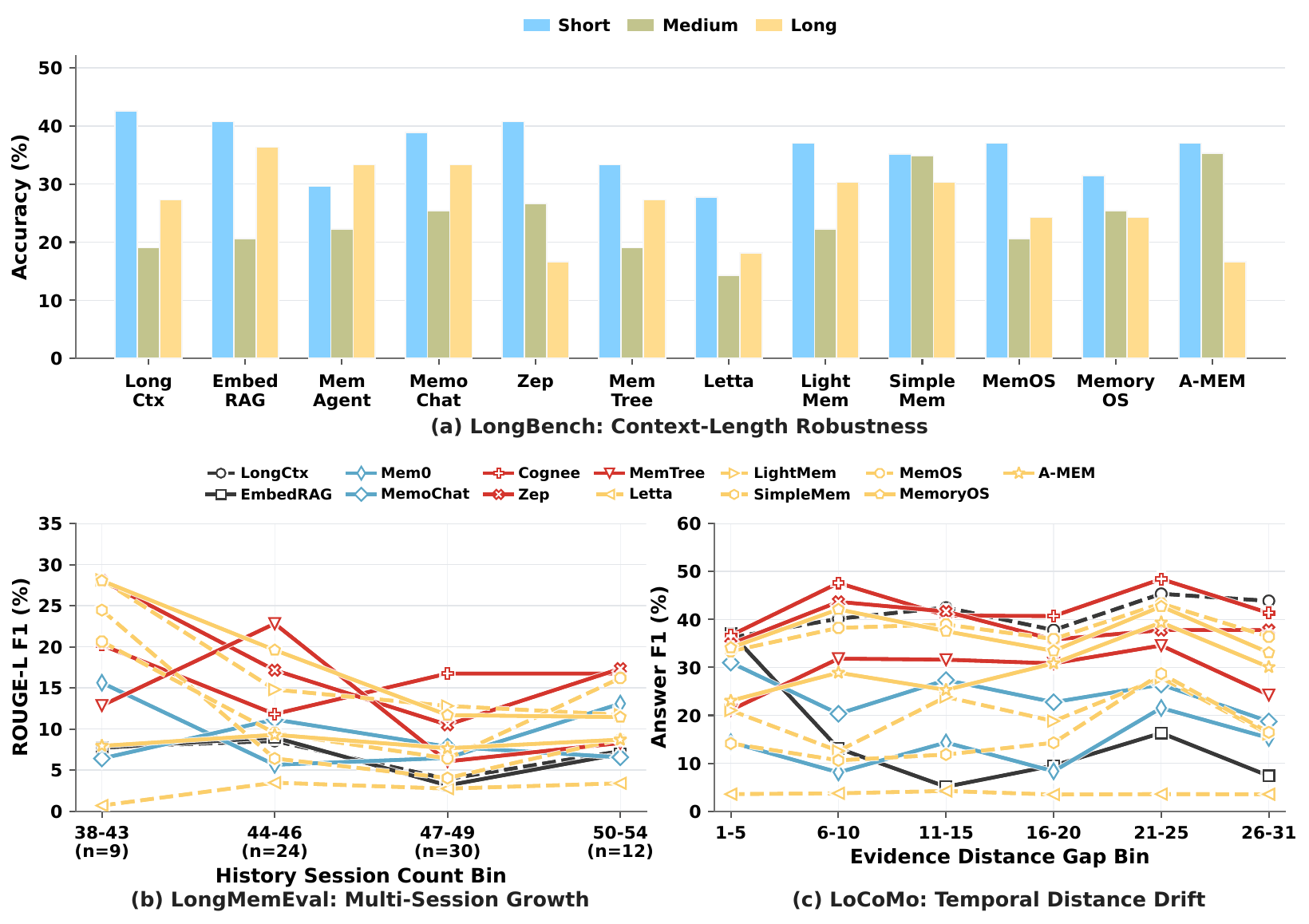}
  \caption{(a) Context-length robustness on \textit{LongBench}; (b) Session-history growth on \textit{LongMemEval}; (c) Temporal evidence-distance drift on \textit{LoCoMo}.}
  \label{fig:RQ4_horizon}
  \vspace{-.75cm}
\end{figure}

\noindent \textbf{O6-(Long-Horizon Evidence Preservation): Memory remains more stable at longer horizons when evidence is organized through explicit relational links or hierarchical consolidation, rather than left as flat text for direct matching.}
As shown in Figure~\ref{fig:RQ4_horizon}, in \textit{LongBench}, \textit{SimpleMem} stays nearly unchanged from the Short to Medium buckets ($35.2$ to $34.9$ \textit{Accuracy}), whereas \textit{Long Context} drops from $42.6$ to $19.0$, indicating that larger prompts alone do not sustain answer quality once long inputs accumulate distractors.
In \textit{LoCoMo}, the contrast is sharper: \textit{Embedding RAG} falls from $37.1$ to $7.4$ \textit{Answer F1} as the evidence gap widens, while graph- or consolidated-memory systems such as \textit{Cognee}, \textit{MemOS}, and \textit{MemoryOS} remain substantially higher across the same bins; \textit{LongMemEval} shows the same advantage for methods that preserve cross-session structure over longer histories.
It indicates that the main difficulty at longer horizons is not memory volume, but whether the representation keeps distant facts connected to the abstractions needed for answering.
More specifically, graph- or temporally organized memory preserves entity--event--time relations for distant facts (e.g., recovering a repeated personal event many sessions earlier), while hierarchical or summary-first organization preserves session-level structure (e.g., first locating the relevant session before resolving a specific local detail) so the \llm can narrow attention before final generation.
Pure long-context prompting and flat dense memory provide neither form of support, and therefore degrade more sharply as the effective horizon grows.

\begin{findingbox}
\textbf{(Horizon-Structured Memory).}
RQ4 indicates that, as the effective memory horizon grows, the main challenge shifts from storing more history to choosing the right abstraction over it:
\emph{(1) Multi-view filtering} helps when long inputs contain many distractors, as in \textit{SimpleMem};
\emph{(2) Relation-aware indexing} helps when supporting facts are separated by many turns or sessions, as in \textit{Cognee} and \textit{Zep};
and  \emph{(3) Coarse-to-fine summarization} helps when the system must first identify the relevant session before resolving a local detail, as in \textit{MemOS} and \textit{MemoryOS}.
\end{findingbox}


\subsection{Memory Operation Cost (RQ5)}
\label{subsec:cost}


\noindent \textbf{Experimental Setting.}
For \textit{``What is the operational cost of each memory system in terms of utility--latency trade-off and cross-workload latency footprint?''}, we evaluate 8 representative memory systems using the unified time-overhead traces recorded by our runner.
We quantify two aspects:
(1) \textit{Utility--latency trade-off}, measured by \textit{Avg.\ Operation Latency/Query} and \textit{Normalized Utility};
and (2) \textit{Cross-workload latency footprint}, measured by \textit{Outlier-Filtered Avg.\ Total Latency/Query}.
For (1), \textit{Avg.\ Operation Latency/Query} is computed as memory construction time plus query time and interpreted as amortized per-query cost for systems with cumulative or bursty writes, while \textit{Normalized Utility} is the mean of six min--max normalized answer-quality metrics from the current \textit{LoCoMo}~\cite{maharana2024locomo} and \textit{LongMemEval}~\cite{wu2024longmemeval} runs.
For (2), we report \textit{Outlier-Filtered Avg.\ Total Latency/Query} on three benchmark.

\begin{figure}[!t]
  \centering
  \includegraphics[width=\linewidth]{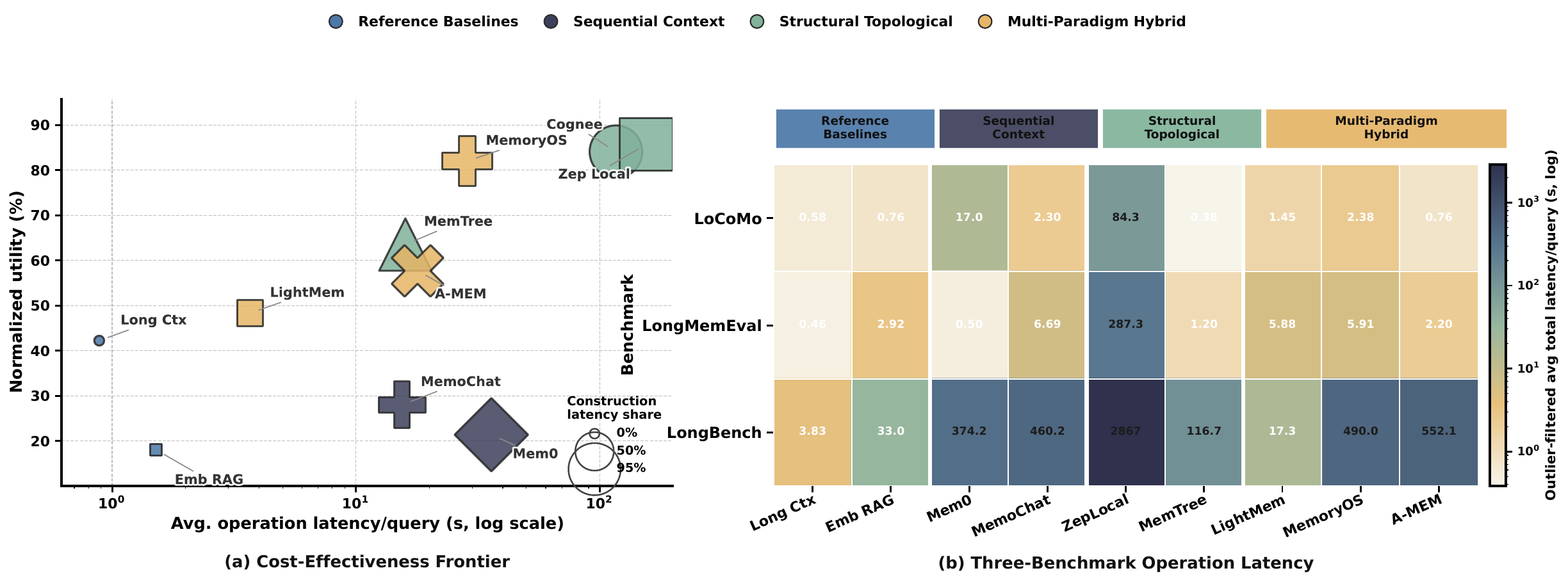}
  \caption{Operation Cost of Memory Systems.}
  \label{fig:RQ5_cost}
  \vspace{-.75cm}
\end{figure}

\noindent \textbf{O7-(Localized Maintenance): The most cost-efficient memory mechanisms are those that localize maintenance to a bounded subset of memory state, whereas mechanisms that repeatedly reorganize a large global state are the least efficient.}
As shown in Figure~\ref{fig:RQ5_cost}, (1) Among memory-augmented systems, LightMem and MemTree occupy the strongest efficiency frontier, with LightMem reaching 48.3 \textit{Normalized Utility} at 3.67\,s \textit{Avg.\ Operation Latency/Query} and MemTree reaching 63.5 at 15.9\,s; both are clearly more efficient than MemoChat (28.0 at 15.4\,s), Mem0 (21.4 at 35.9\,s), and A-MEM (57.7 at 17.9\,s);
(2) Higher-utility structured systems move markedly to the expensive side: MemoryOS reaches 82.0 \textit{Normalized Utility} only at 28.6\,s, while Cognee and Zep exceed 84 utility only after 116.5\,s and 155.1\,s;
(3) The workload-specific latency view sharpens the same separation on \textit{LongBench}: LightMem remains at 17.3\,s and MemTree at 116.7\,s, whereas Mem0, MemoChat, MemoryOS, and A-MEM rise to 374.2, 460.2, 490.0, and 552.1\,s, respectively.
It indicates that operational efficiency is governed less by whether a system uses structure than by how widely each write propagates through that structure.
Specifically, (1) segmented compression and bounded hybrid retrieval keep LightMem close to the low-cost regime;
(2) path-local tree aggregation allows MemTree to preserve substantially more utility without global refresh;
and (3) graph-wide consolidation, multi-store synchronization, or repeated whole-memory rewriting yield stronger organization but impose the heaviest operational cost as memory grows.

\begin{findingbox}
\textbf{(Operational Scaling Rule).}
RQ5 shows that efficiency is governed by maintenance scope rather than structure alone.
(1) Localized update and search yield the strongest cost--utility balance, as in LightMem and MemTree;
(2) Richer organization helps only when its upkeep avoids broad recomputation; otherwise overhead offsets its gains, as in Cognee and MemoryOS;
(3) Under long-context workloads, whole-memory coordination becomes the dominant cost driver.
\end{findingbox}

\section{Fine-Grained Component Comparison}
\label{sec:finegrained}

To understand the root causes behind end-to-end performance differences, we decompose agent memory systems into four fundamental modules.
By systematically generating controlled variants that modify one module at a time, we evaluate the contribution of each module to overall system performance.

\subsection{Memory Representation and Storage (M1)}
\label{subsec:representation}



\begin{table}[!t]
\caption{Ablation of Representation and Storage Mechanisms.}
\label{tab:m1}
\centering
\footnotesize
\setlength{\tabcolsep}{4.0pt}
\resizebox{\linewidth}{!}{%
\begin{tabular}{llcccc}
\toprule
\multicolumn{2}{c}{\textbf{Method Variant}} & \multicolumn{2}{c}{\textbf{LoCoMo}} & \multicolumn{2}{c}{\textbf{LongMemEval}} \\
\cmidrule(lr){3-4} \cmidrule(lr){5-6}
 & & \textbf{EM} & \textbf{Ans.\ F1} & \textbf{Substr.\ EM} & \textbf{ROUGE-L F1} \\
\midrule
\multirow{3}{*}{\textbf{LightMem}} & User-Only Raw & \textbf{24.2} & \textbf{38.9} & \textbf{26.0} & \textbf{31.4} \\
 & User-Only Summary & 8.5 & 15.6 & 11.7 & 17.4 \\
 & User-Only Compressed & 23.6 & 38.6 & 10.7 & 19.1 \\
\midrule
\multirow{2}{*}{\textbf{MemTree}} & Flat-biased & 18.2 & 30.7 & 23.0 & 29.9 \\
 & Deeper Tree & \textbf{18.7} & \textbf{31.2} & \textbf{23.3} & \textbf{30.9} \\
\midrule
\multirow{2}{*}{\textbf{Mem0}} & Default & \textbf{3.2} & 6.2 & \textbf{9.3} & \textbf{16.5} \\
 & Graph Store & 3.0 & \textbf{6.5} & 8.3 & 15.9 \\
\bottomrule
\end{tabular}
}
\end{table}

\noindent \textbf{Experimental Setting.}
For \textit{``How do memory abstraction level and structural organization affect factual fidelity and downstream reasoning effectiveness?''}, we evaluate three representation-focused variants:
(1) \textit{LightMem}~\cite{LightMem} compares \textit{User-Only Raw}, which stores verbatim user utterances, \textit{User-Only Summary}, which rewrites each session into an LLM-generated abstractive summary, and \textit{User-Only Compressed}, which removes filler and redundant tokens while preserving the original phrasing and factual content;
(2) \textit{MemTree}~\cite{MemTree} compares a shallow \textit{Flat-biased Setting} with a \textit{Deeper Tree Setting} to examine how hierarchical text organization affects memory fidelity.
We assess the trade-off between fine-grained fact preservation and multi-step reasoning using \textit{LoCoMo} for compositional reasoning and \textit{LongMemEval} for multi-session factual retrieval, reporting exactness- and overlap-based metrics (e.g., \textit{EM} and \textit{ROUGE-L F1}).

\noindent \textbf{O8-(Content Fidelity): Retaining the original conversational content is more important than increasing abstraction or hierarchy for sustaining both factual recall and reasoning quality.}
Table~\ref{tab:m1} shows that \textit{LightMem} \textit{User-Only Raw} achieves the best result on all four metrics, while \textit{User-Only Compressed} remains close on \textit{LoCoMo} (\textit{Ans.\ F1}: 38.6 vs.\ 38.9; \textit{EM}: 23.6 vs.\ 24.2) but drops sharply on \textit{LongMemEval} (\textit{Substring EM}: 10.7 vs.\ 26.0), \textit{User-Only Summary} is substantially weaker on both benchmarks, and the deeper \textit{MemTree} setting provides only modest gains over the flat setting.
It indicates that the main performance boundary is the amount of recoverable evidence the representation preserves, rather than whether it applies stronger abstraction or a deeper structure.
In particular, (1) Raw text is most effective when the task requires recovering exact session-level details (e.g., recalling a title such as \textit{``Nu, pogodi!''}); (2) Light compression can still support compositional reasoning when the main meaning is preserved, but becomes unreliable for exact detail matching (e.g., relating two earlier events while missing the precise date or name); and (3) Deeper hierarchy can improve organization, but cannot restore information removed during representation (e.g., a parent node helps navigate related sessions, but not recover omitted details).

\begin{findingbox}
\textbf{(Representation Granularity).}
M1 shows that preserving usable evidence matters more than making memory more compact or more structured.
(1) \emph{High-retention forms} best support exact detail recovery, as in \textit{LightMem} \textit{User-Only Raw}; (2) \emph{Light compression} can preserve reasoning, but weakens exact matching, as in \textit{LightMem} \textit{User-Only Compressed}; (3) \emph{Hierarchy} mainly improves access, but cannot restore removed content, as reflected by the \textit{MemTree (Deeper Tree)} variant.
\end{findingbox}

\subsection{Memory Extraction (M2)}
\label{subsec:extraction}

\begin{table}[!t]
\caption{Ablation of Memory Extraction Strategies.}
\label{tab:m2_extraction_table}
\centering
\footnotesize
\setlength{\tabcolsep}{4.0pt}
\resizebox{\linewidth}{!}{%
\begin{tabular}{llcccc}
\toprule
\multicolumn{2}{c}{\textbf{Method Variant}} & \multicolumn{2}{c}{\textbf{LoCoMo}} & \multicolumn{2}{c}{\textbf{LongMemEval}} \\
\cmidrule(lr){3-4} \cmidrule(lr){5-6}
 & & \textbf{EM} & \textbf{Ans.\ F1} & \textbf{Substr.\ EM} & \textbf{ROUGE-L F1} \\
\midrule
\multirow{2}{*}{\textbf{MemoChat}} & Heuristic Topic & \textbf{23.0} & 33.5 & \textbf{10.7} & \textbf{18.6} \\
 & LLM Topic & 22.5 & \textbf{34.4} & 7.3 & 15.9 \\
\midrule
\multirow{2}{*}{\textbf{MemOS}} & Fast Memorize & \textbf{25.5} & \textbf{40.8} & 20.7 & 26.1 \\
 & Fine Memorize & 2.5 & 5.0 & \textbf{22.3} & \textbf{30.2} \\
\midrule
\multirow{2}{*}{\textbf{LightMem}} & User-Only Raw & 24.2 & 38.9 & \textbf{26.0} & \textbf{31.4} \\
 & Hybrid Raw & \textbf{25.5} & \textbf{39.7} & 25.3 & \textbf{31.4} \\
\bottomrule
\end{tabular}
}
\end{table}


\noindent \textbf{Experimental Setting.}
For \textit{``How do write-time extraction choices affect factual fidelity and downstream reasoning effectiveness?''}, we compare extraction-related variants in three groups:
\textit{(1) MemoChat}~\cite{Memochat}, which contrasts \textit{Heuristic Topic} and \textit{LLM Topic} segmentation;
\textit{(2) MemOS}~\cite{MemOS}, which contrasts \textit{Fast Memorize} and \textit{Fine Memorize} on the same \texttt{tree\_text} backend;
and \textit{(3) LightMem}~\cite{LightMem}, which contrasts \textit{User-Only Raw} and \textit{Hybrid Raw} by extracting raw memory from user turns only or from both user and assistant turns.
We evaluate \textit{LongMemEval} for multi-session factual retrieval fidelity and \textit{LoCoMo} for downstream multi-step reasoning, reporting exactness- and overlap-based measures (e.g., \textit{EM} and \textit{ROUGE-L F1}).

\noindent \textbf{O9-(Coverage-Preserving Extraction): Coverage-preserving write-time extraction provides the most stable balance between factual retrieval and downstream reasoning.}
As shown in Table~\ref{tab:m2_extraction_table}, \textit{MemoChat} \textit{Heuristic Topic} improves LongMemEval over \textit{LLM Topic} (10.7 vs.\ 7.3 \textit{Substr.\ EM}; 18.6 vs.\ 15.9 \textit{ROUGE-L F1}) while keeping LoCoMo nearly unchanged (23.0/33.5 vs.\ 22.5/34.4 \textit{EM}/\textit{Ans.\ F1}), \textit{MemOS} \textit{Fast Memorize} far exceeds \textit{Fine Memorize} on LoCoMo (25.5 vs.\ 2.5 \textit{EM}; 40.8 vs.\ 5.0 \textit{Ans.\ F1}) despite lower LongMemEval scores (20.7 vs.\ 22.3 \textit{Substr.\ EM}; 26.1 vs.\ 30.2 \textit{ROUGE-L F1}), and \textit{LightMem} \textit{Hybrid Raw} slightly improves LoCoMo over \textit{User-Only Raw} (25.5 vs.\ 24.2 \textit{EM}; 39.7 vs.\ 38.9 \textit{Ans.\ F1}) with nearly unchanged LongMemEval results (25.3 vs.\ 26.0 \textit{Substr.\ EM}; both 31.4 \textit{ROUGE-L F1}).
These results suggest that broader, less selective extraction better preserves the context needed for downstream answerability, even when more selective extraction yields modest gains on lexical factual retrieval.
In particular, (1) conservative topic grouping is less likely to split a sustained thread or isolate a brief aside (e.g., a one-off hobby mention); (2) lighter memorization is more likely to retain details that later need to be combined for reasoning; 
and (3) including both user and assistant turns can preserve clarifying cues that user-only extraction may miss (e.g., a date or refined phrasing).

\begin{findingbox}
\textbf{(Late Filtering Principle).}
M2 suggests that memory extraction should preserve context at write time rather than aggressively filter details: 
(1) Coarser segmentation helps thread-spanning questions by keeping related cues together;
(2) Limited rewriting supports compositional reasoning by retaining details that matter only when combined later;
(3) Storing both user and assistant turns helps clarification-heavy dialogues by preserving refined formulations for later access.
\end{findingbox}

\subsection{Memory Retrieval and Routing (M3)}
\label{subsec:retrieval}




\begin{table}[!t]
\caption{Ablation of Retrieval and Routing Mechanisms.}
\vspace{-.25cm}
\label{tab:m3_retrieval}
\centering
\footnotesize
\setlength{\tabcolsep}{4.0pt}
\resizebox{\linewidth}{!}{%
\begin{tabular}{llcccc}
\toprule
\multicolumn{2}{c}{\textbf{Method Variant}} & \multicolumn{2}{c}{\textbf{LoCoMo}} & \multicolumn{2}{c}{\textbf{LongMemEval}} \\
\cmidrule(lr){3-4} \cmidrule(lr){5-6}
 & & \textbf{Ans.\ F1} & \textbf{Recall} & \textbf{Substr.\ EM} & \textbf{ROUGE-L F1} \\
\midrule
\multirow{2}{*}{\textbf{A-MEM}} & Hybrid-Balanced & \textbf{24.6} & 49.9 & \textbf{27.5} & \textbf{25.9} \\
 & Hybrid Sparse-Leaning & 23.0 & 44.3 & 24.3 & 22.8 \\
\midrule
\multirow{3}{*}{\textbf{SimpleMem}} & No Planning & 18.7 & 86.4 & 17.0 & 22.9 \\
 & Planning Only & \textbf{20.7} & \textbf{90.6} & \textbf{21.7} & \textbf{27.9} \\
 & Planning + Reflect & 20.0 & 88.6 & 21.3 & 26.1 \\
\bottomrule
\end{tabular}
}
\end{table}

\noindent \textbf{Experimental Setting.}
For \textit{``How do retrieval fusion and reasoning-mediated routing affect retrieval relevance and provenance-sensitive precision?''}, we compare variants in two groups:
(1) \textit{A-MEM} under \textit{Hybrid-Balanced}, which uses a moderate dense--sparse fusion, versus \textit{Hybrid Sparse-Leaning}, which increases the sparse contribution; and (2) \textit{SimpleMem} under \textit{No Planning}, which retrieves directly, versus \textit{Planning Only}, which adds an explicit planning step, and \textit{Planning + Reflect}, which further introduces a lightweight reflection stage.
We evaluate on \textit{LongMemEval} to measure scattered-history retrieval relevance, and on \textit{LoCoMo} to assess provenance-sensitive memory access and supporting-memory identification, reporting overlap-based measures (e.g., \textit{Substr. EM}, \textit{ROUGE-L F1}).

\noindent \textbf{O10-(Planning and Fusion): Explicit planning and balanced retrieval fusion provide the strongest improvement in retrieval effectiveness.}
As shown in Table~\ref{tab:m3_retrieval}, \textit{A-MEM} achieves its best performance with \textit{Hybrid-Balanced}, reaching 24.6 \textit{Ans.\ F1} and 27.5 \textit{Substr.\ EM}, compared with 23.0 and 24.3 under \textit{Hybrid Sparse-Leaning}, while \textit{SimpleMem} achieves its best performance with \textit{Planning Only}, reaching 20.7 \textit{Ans.\ F1}, 90.6 \textit{Strict Rec.}, 21.7 \textit{Substr.\ EM}, and 27.9 \textit{ROUGE-L F1}, above both \textit{No Planning} and \textit{Planning + Reflect}.
These results indicate that stronger retrieval and routing performance comes from adding useful structure, rather than from simply increasing sparse matching or extra reasoning steps.
In particular, (1) moderate fusion appears more effective than sparse-leaning fusion for preserving both answer quality and relevance (e.g., semantically related but lexically varied facts); (2) explicit planning consistently improves over direct retrieval (e.g., multi-constraint memory queries); and (3) adding reflection on top of planning does not yield further gains, suggesting that extra deliberation may weaken rather than improve routing decisions.

\begin{findingbox}
\textbf{(Retrieval Strategy Guidance).}
M3 indicates that retrieval quality improves most from targeted structure rather than added complexity:
(1) moderate hybrid fusion is preferable when evidence is semantically related but lexically diverse;
(2) lightweight planning is effective for constrained memory lookup;
and (3) once a route is already specified, extra reflection brings limited benefit and mainly adds overhead.
\end{findingbox}

\subsection{Memory Maintenance (M4)}
\label{subsec:maintenance}

\begin{figure}[!t]
  \centering
  \includegraphics[width=\linewidth]{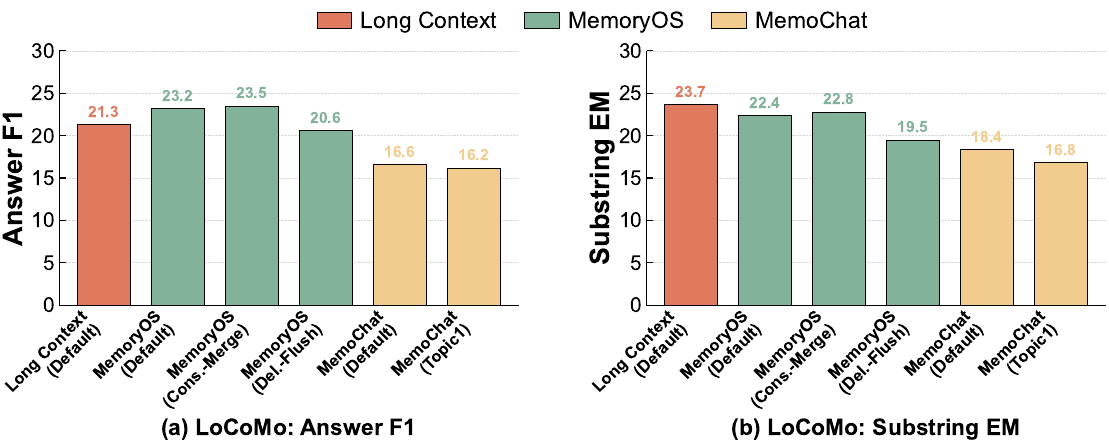}
  \vspace{-.65cm}
  \caption {Ablation of Maintenance Strategies.}
  \label{fig:m4_maintenance}
  \vspace{-.5cm}
\end{figure}


\noindent \textbf{Experimental Setting.}
For \textit{``How do consolidation aggressiveness, flush timing, and summary granularity affect update correctness and long-horizon memory consistency?''}, we compare maintenance-relevant variants in two groups:
\textit{(1) MemoChat} under default multi-topic consolidation versus \textit{Topic1}, which forces each window into a single-topic summary;
and \textit{(2) MemoryOS} under default immediate consolidation versus \textit{Delayed-Flush}, which enlarges the short-term buffer before backend writes, and \textit{Conservative-Merge}, which raises the topic-similarity threshold for stricter assimilation.
We evaluate on \textit{LoCoMo} to assess whether these maintenance choices preserve updated facts and coherent memory use over extended contexts.

\begin{sloppypar}
\noindent \textbf{O11-(Conservative Consolidation): Conservative consolidation is more effective than delayed flushing or overly coarse summarization for maintaining answer-relevant memory.}
As shown in Figure~\ref{fig:m4_maintenance}, the stricter-merge variant, \textit{MemoryOS (Conservative-Merge)}, improves over default \textit{MemoryOS} from 23.2 to 23.5 in \textit{Ans.\ F1} and from 22.4 to 22.8 in \textit{Substr.\ EM}, whereas delaying flushes lowers the same system to 20.6/19.5, and forcing single-topic summaries in \textit{MemoChat} also underperforms its default setting at 16.2/16.8 versus 16.6/18.4; \textit{Long Context} remains highest on \textit{Substr.\ EM} at 23.7.
It indicates that maintenance is most effective when it selectively consolidates evidence without either leaving it unresolved or compressing it too aggressively, while raw context still better preserves exact phrasing.
In particular, (1) conservative merging can retain related details for later recomposition (e.g., dispersed hobby mentions);
(2) delayed flushing leaves more evidence unresolved before retrieval (e.g., activity split across turns).
\end{sloppypar}

\begin{findingbox}
\textbf{(Maintenance Design Principle).} M4 suggests that memory maintenance works best under a balanced update regime: 
(1) conservative integration preserves cross-turn linkages for long-horizon reasoning; 
(2) delayed flushing leaves recent evidence fragmented at query time; 
and (3) overly coarse summarization obscures sparse but useful cues.
\end{findingbox}

\section{Conclusion}
\label{sec:conclusion}

We present a comprehensive review of existing agent memory systems from a data management perspective. We conduct thorough end-to-end performance evaluation of typical agent memory systems and explore their suitable application scenarios. Additionally, we delve into the impact of the individual building blocks by constructing multiple memory module variants, thereby identifying the most effective methods for representation, extraction, routing, and maintenance, as well as the most influential factors governing operational costs and long-horizon stability. Finally, we summarize the findings and present guidance for users in selecting suitable memory architectures, alongside outlining promising research directions. We will also release the testbed and evaluation framework.

\clearpage

\balance

\bibliographystyle{ACM-Reference-Format}
\bibliography{sample}

@article{LightMem,
  author       = {Jizhan Fang and
                  Xinle Deng and
                  Haoming Xu and
                  Ziyan Jiang and
                  Yuqi Tang and
                  Ziwen Xu and
                  Shumin Deng and
                  Yunzhi Yao and
                  Mengru Wang and
                  Shuofei Qiao and
                  Huajun Chen and
                  Ningyu Zhang},
  title        = {LightMem: Lightweight and Efficient Memory-Augmented Generation},
  journal      = {CoRR},
  volume       = {abs/2510.18866},
  year         = {2025},
  url          = {https://doi.org/10.48550/arXiv.2510.18866},
  doi          = {10.48550/ARXIV.2510.18866},
  eprinttype   = {arXiv},
  eprint       = {2510.18866},
  bibsource    = {dblp computer science bibliography, https://dblp.org}
}

@inproceedings{MemTree,
  author       = {Alireza Rezazadeh and
                  Zichao Li and
                  Wei Wei and
                  Yujia Bao},
  title        = {From Isolated Conversations to Hierarchical Schemas: Dynamic Tree
                  Memory Representation for LLMs},
  booktitle    = {The Thirteenth International Conference on Learning Representations,
                  {ICLR} 2025, Singapore, April 24-28, 2025},
  publisher    = {OpenReview.net},
  year         = {2025},
  url          = {https://openreview.net/forum?id=moXtEmCleY},
  bibsource    = {dblp computer science bibliography, https://dblp.org}
}

@article{Mem0,
  author       = {Prateek Chhikara and
                  Dev Khant and
                  Saket Aryan and
                  Taranjeet Singh and
                  Deshraj Yadav},
  title        = {Mem0: Building Production-Ready {AI} Agents with Scalable Long-Term
                  Memory},
  journal      = {CoRR},
  volume       = {abs/2504.19413},
  year         = {2025},
  url          = {https://doi.org/10.48550/arXiv.2504.19413},
  doi          = {10.48550/ARXIV.2504.19413},
  eprinttype   = {arXiv},
  eprint       = {2504.19413},
  bibsource    = {dblp computer science bibliography, https://dblp.org}
}

@article{Memochat,
  author       = {Junru Lu and
                  Siyu An and
                  Mingbao Lin and
                  Gabriele Pergola and
                  Yulan He and
                  Di Yin and
                  Xing Sun and
                  Yunsheng Wu},
  title        = {MemoChat: Tuning LLMs to Use Memos for Consistent Long-Range Open-Domain
                  Conversation},
  journal      = {CoRR},
  volume       = {abs/2308.08239},
  year         = {2023},
  url          = {https://doi.org/10.48550/arXiv.2308.08239},
  doi          = {10.48550/ARXIV.2308.08239},
  eprinttype   = {arXiv},
  eprint       = {2308.08239},
  bibsource    = {dblp computer science bibliography, https://dblp.org}
}

@article{MemOS,
  author       = {Zhiyu Li and
                  Shichao Song and
                  Chenyang Xi and
                  Hanyu Wang and
                  Chen Tang and
                  Simin Niu and
                  Ding Chen and
                  Jiawei Yang and
                  Chunyu Li and
                  Qingchen Yu and
                  Jihao Zhao and
                  Yezhaohui Wang and
                  Peng Liu and
                  Zehao Lin and
                  Pengyuan Wang and
                  Jiahao Huo and
                  Tianyi Chen and
                  Kai Chen and
                  Kehang Li and
                  Zhen Tao and
                  Junpeng Ren and
                  Huayi Lai and
                  Hao Wu and
                  Bo Tang and
                  Zhenren Wang and
                  Zhaoxin Fan and
                  Ningyu Zhang and
                  Linfeng Zhang and
                  Junchi Yan and
                  Mingchuan Yang and
                  Tong Xu and
                  Wei Xu and
                  Huajun Chen and
                  Haofeng Wang and
                  Hongkang Yang and
                  Wentao Zhang and
                  Zhi{-}Qin John Xu and
                  Siheng Chen and
                  Feiyu Xiong},
  title        = {MemOS: {A} Memory {OS} for {AI} System},
  journal      = {CoRR},
  volume       = {abs/2507.03724},
  year         = {2025},
  url          = {https://doi.org/10.48550/arXiv.2507.03724},
  doi          = {10.48550/ARXIV.2507.03724},
  eprinttype   = {arXiv},
  eprint       = {2507.03724},
  bibsource    = {dblp computer science bibliography, https://dblp.org}
}

@article{llm4db,
  author       = {Guoliang Li and
                  Xuanhe Zhou and
                  Xinyang Zhao},
  title        = {{LLM} for Data Management},
  journal      = {Proc. {VLDB} Endow.},
  volume       = {17},
  number       = {12},
  pages        = {4213--4216},
  year         = {2024}
}

@article{RAG,
  author       = {Yunfan Gao and
                  Yun Xiong and
                  Xinyu Gao and
                  Kangxiang Jia and
                  Jinliu Pan and
                  Yuxi Bi and
                  Yi Dai and
                  Jiawei Sun and
                  Qianyu Guo and
                  Meng Wang and
                  Haofen Wang},
  title        = {Retrieval-Augmented Generation for Large Language Models: {A} Survey},
  journal      = {CoRR},
  volume       = {abs/2312.10997},
  year         = {2023}
}

@article{llm,
  author       = {Wayne Xin Zhao and
                  Kun Zhou and
                  Junyi Li and
                  Tianyi Tang and
                  Xiaolei Wang and
                  Yupeng Hou and
                  Yingqian Min and
                  Beichen Zhang and
                  Junjie Zhang and
                  Zican Dong and
                  Yifan Du and
                  Chen Yang and
                  Yushuo Chen and
                  Zhipeng Chen and
                  Jinhao Jiang and
                  Ruiyang Ren and
                  Yifan Li and
                  Xinyu Tang and
                  Zikang Liu and
                  Peiyu Liu and
                  Jian{-}Yun Nie and
                  Ji{-}Rong Wen},
  title        = {A Survey of Large Language Models},
  journal      = {CoRR},
  volume       = {abs/2303.18223},
  year         = {2023}
}

@article{zhou2026dbcooker,
  author       = {Wei Zhou and Xuanhe Zhou and Qikang He and Guoliang Li and Bingsheng He and Quanqing Xu and Fan Wu},
  title        = {Automating
Database-Native Function Code Synthesis with LLMs},
  journal      = {Proc. {ACM} Manag. Data},
  volume       = {3},
  number       = {4},
  pages        = {141:1--141:26},
  year         = {2026}
}

@article{luo2026data,
  title={Data Agents: Levels, State of the Art, and Open Problems},
  author={Luo, Yuyu and Li, Guoliang and Fan, Ju and Tang, Nan},
  journal={arXiv preprint arXiv:2602.04261},
  note={SIGMOD 2026 Tutorial},
  year={2026}
}

@article{khan2026rag,
  title={Retrieval-augmented Generation ({RAG}): What is There for Data Management Researchers?},
  author={Khan, Arijit and Luo, Yuyu and Zhang, Wenjie and Zhou, Mingjie and Zhou, Xiaofang},
  journal={ACM SIGMOD Record},
  volume={54},
  number={4},
  year={2025},
  publisher={ACM}
}

@inproceedings{liu2026supporting,
  title={Supporting Our {AI} Overlords: Redesigning Data Systems to be Agent-First},
  author={Liu, Shu and Ponnapalli, Soujanya and Shankar, Shreya and Zeighami, Sepanta and Zhu, Alan and Agarwal, Shubham and Chen, Ruiqi and Suwito, Samion and Yuan, Shuo and Stoica, Ion and Zaharia, Matei and Cheung, Alvin and Crooks, Natacha and Gonzalez, Joseph E. and Parameswaran, Aditya G.},
  booktitle={Proceedings of the 16th Annual Conference on Innovative Data Systems Research (CIDR)},
  year={2026}
}

@article{kang2025bigvectorbench,
  title={{BigVectorBench}: Heterogeneous Data Embedding and Compound Queries are Essential in Evaluating Vector Databases},
  author={Kang, Guozhang and Ge, Zhenying and Hu, Jie and Zhang, Xinyuan and Wang, Li and Zhan, Jianfeng},
  journal={Proceedings of the VLDB Endowment},
  volume={18},
  number={6},
  pages={1536--1549},
  year={2025}
}

@inproceedings{caminal2025filtered,
  title={Filtered Vector Search: State-of-the-art and Research Opportunities},
  author={Caminal, Liana and others},
  booktitle={Proceedings of the VLDB Endowment},
  volume={18},
  pages={5488--5491},
  year={2025}
}

@article{packer2023memgpt,
  title={{MemGPT}: Towards {LLMs} as Operating Systems},
  author={Packer, Charles and Fang, Vivian and Patil, Shishir G. and Lin, Kevin and Wooders, Sarah and Gonzalez, Joseph E.},
  journal={arXiv preprint arXiv:2310.08560},
  year={2023}
}

@article{chhikara2025mem0,
  title={{Mem0}: Building Production-Ready {AI} Agents with Scalable Long-Term Memory},
  author={Chhikara, Prateek and Khant, Dev and Aryan, Saket and Singh, Taranjeet and Yadav, Deshraj},
  journal={arXiv preprint arXiv:2504.19413},
  year={2025}
}

@article{xu2025amem,
  title={{A-MEM}: Agentic Memory for {LLM} Agents},
  author={Xu, Wujiang and others},
  journal={arXiv preprint arXiv:2502.12110},
  year={2025}
}

@article{rasmussen2025zep,
  title={Zep: A Temporal Knowledge Graph Architecture for Agent Memory},
  author={Rasmussen, Preston and Paliychuk, Pavel and Beauvais, Travis and Ryan, Jesse},
  journal={arXiv preprint arXiv:2501.13956},
  year={2025}
}

@article{hu2025memory,
  title={Memory in the Age of {AI} Agents},
  author={Hu, Yifan and Liu, Siyin and Yue, Yifei and Zhang, Guoqiang and Liu, Benyou and Zhu, Fengbin and Lin, Jingkuan and others},
  journal={arXiv preprint arXiv:2512.13564},
  year={2025}
}

@article{yang2026graph,
  title={Graph-based Agent Memory: Taxonomy, Techniques, and Applications},
  author={Yang, Chao and Zhou, Chuan and Xiao, Yanghua and Dong, Shuai and Zhuang, Liang and others},
  journal={arXiv preprint arXiv:2602.05665},
  year={2026}
}

@inproceedings{maharana2024locomo,
  title={Evaluating Very Long-Term Conversational Memory of {LLM} Agents},
  author={Maharana, Adyasha and Lee, Dong-Ho and Turishcheva, Sergey and Nham, Kezhen and Jandaghi, Golnaz and Pujara, Jay and Ren, Xiang},
  booktitle={Proceedings of the 62nd Annual Meeting of the Association for Computational Linguistics (ACL)},
  year={2024}
}

@article{wu2024longmemeval,
  title={{LongMemEval}: Benchmarking Chat Assistants on Long-Term Interactive Memory},
  author={Wu, Di and Wang, Hongwei and Yu, Wenhao and Zhang, Yuwei and Chang, Kai-Wei},
  journal={arXiv preprint arXiv:2410.10813},
  year={2024}
}

@inproceedings{memoryagentbench2026,
  title={Evaluating Memory in {LLM} Agents via Incremental Multi-Turn Interactions},
  author={{MemoryAgentBench Team}},
  booktitle={Fourteenth International Conference on Learning Representations (ICLR)},
  year={2026}
}

@misc{openai2026agents,
  title={Context Engineering for Personalization -- State Management with Long-Term Memory Notes using {OpenAI} Agents {SDK}},
  author={{OpenAI}},
  year={2026},
  howpublished={\url{https://developers.openai.com/cookbook/examples/agents_sdk/context_personalization/}}
}

@misc{anthropic2025context,
  title={Effective context engineering for {AI} agents},
  author={{Anthropic Engineering}},
  year={2025},
  howpublished={\url{https://www.anthropic.com/engineering/effective-context-engineering-for-ai-agents}}
}

@misc{microsoft2025copilot,
  title={Introducing Copilot Memory: A More Productive and Personalized {AI}},
  author={{Microsoft}},
  year={2025},
  howpublished={\url{https://techcommunity.microsoft.com/blog/microsoft365copilotblog/introducing-copilot-memory}}
}

@misc{google2025adk,
  title={Memory -- Agent Development Kit ({ADK})},
  author={{Google}},
  year={2025},
  howpublished={\url{https://google.github.io/adk-docs/sessions/memory/}}
}

@inproceedings{singh2024personal,
  title     = {Personal Large Language Model Agents: A Case Study on Tailored Travel Planning},
  author    = {Singh, Harmanpreet and Verma, Nikhil and Wang, Yixiao and Bharadwaj, Manasa and Fashandi, Homa and Ferreira, Kevin and Lee, Chul},
  booktitle = {Proceedings of the 2024 Conference on Empirical Methods in Natural Language Processing: Industry Track},
  pages     = {486--514},
  year      = {2024},
  address   = {Miami, Florida, US},
  publisher = {Association for Computational Linguistics},
  doi       = {10.18653/v1/2024.emnlp-industry.37}
}

@inproceedings{memorybank,
  author       = {Wanjun Zhong and
                  Lianghong Guo and
                  Qiqi Gao and
                  He Ye and
                  Yanlin Wang},
  title        = {MemoryBank: Enhancing Large Language Models with Long-Term Memory},
  booktitle    = {{AAAI}},
  pages        = {19724--19731},
  publisher    = {{AAAI} Press},
  year         = {2024}
}

@inproceedings{membench,
  author       = {Haoran Tan and
                  Zeyu Zhang and
                  Chen Ma and
                  Xu Chen and
                  Quanyu Dai and
                  Zhenhua Dong},
  title        = {MemBench: Towards More Comprehensive Evaluation on the Memory of LLM-based
                  Agents},
  booktitle    = {{ACL} (Findings)},
  series       = {Findings of {ACL}},
  pages        = {19336--19352},
  publisher    = {Association for Computational Linguistics},
  year         = {2025}
}

@article{zheng2025lifelong,
  title={Lifelong Learning of Large Language Model based Agents: A Roadmap},
  author={Zheng, Junhao and Shi, Chengming and Cai, Xidi and Li, Qiuke and Zhang, Duzhen and Li, Chenxing and Yu, Dong and Ma, Qianli},
  journal={IEEE Transactions on Pattern Analysis and Machine Intelligence},
  year={2025}
}

@article{du2026memory,
  title={Memory for Autonomous LLM Agents: Mechanisms, Evaluation, and Emerging Frontiers},
  author={Du, Pengfei},
  journal={arXiv preprint arXiv:2603.07670},
  year={2026}
}

@article{wu2026memoryeab,
  title={Memory in the {LLM} Era: Modular Architectures and Strategies in a Unified Framework},
  author={Wu, Yanchen and Lin, Tenghui and Zhou, Yingli and Zhang, Fangyuan and Guo, Qintian and Zhou, Xun and Wang, Sibo and Liu, Xilin and Ma, Yuchi and Fang, Yixiang},
  journal={Proceedings of the VLDB Endowment},
  year={2026}
}

@article{zhang2024survey,
  title={A Survey on the Memory Mechanism of Large Language Model based Agents},
  author={Zhang, Zeyu and Bo, Xiaohe and Ma, Chen and Li, Rui and Chen, Xu and Dai, Quanyu and Zhu, Jieming and Dong, Zhenhua and Wen, Ji-Rong},
  journal={ACM Transactions on Information Systems},
  year={2025}
}

@article{tang2026survey,
  title={{LLM} Agent Memory: A Survey from a Unified Representation},
  author={Tang, Zhiwei and others},
  journal={arXiv preprint arXiv:2603.0359},
  year={2026}
}

@inproceedings{DBLP:conf/acl/BaiLZL0HDLZHDTL24,
  author       = {Yushi Bai and
                  Xin Lv and
                  Jiajie Zhang and
                  Hongchang Lyu and
                  Jiankai Tang and
                  Zhidian Huang and
                  Zhengxiao Du and
                  Xiao Liu and
                  Aohan Zeng and
                  Lei Hou and
                  Yuxiao Dong and
                  Jie Tang and
                  Juanzi Li},
  title        = {LongBench: {A} Bilingual, Multitask Benchmark for Long Context Understanding},
  booktitle    = {{ACL} {(1)}},
  pages        = {3119--3137},
  publisher    = {Association for Computational Linguistics},
  year         = {2024}
}

@article{DBLP:journals/corr/abs-2505-11942,
  author       = {Junhao Zheng and
                  Xidi Cai and
                  Qiuke Li and
                  Duzhen Zhang and
                  Zhong{-}Zhi Li and
                  Yingying Zhang and
                  Le Song and
                  Qianli Ma},
  title        = {LifelongAgentBench: Evaluating {LLM} Agents as Lifelong Learners},
  journal      = {CoRR},
  volume       = {abs/2505.11942},
  year         = {2025}
}

@article{simplemem,
  author       = {Jiaqi Liu and
                  Yaofeng Su and
                  Peng Xia and
                  Siwei Han and
                  Zeyu Zheng and
                  Cihang Xie and
                  Mingyu Ding and
                  Huaxiu Yao},
  title        = {SimpleMem: Efficient Lifelong Memory for {LLM} Agents},
  journal      = {CoRR},
  volume       = {abs/2601.02553},
  year         = {2026}
}

@inproceedings{memoryos,
  author       = {Jiazheng Kang and
                  Mingming Ji and
                  Zhe Zhao and
                  Ting Bai},
  title        = {Memory {OS} of {AI} Agent},
  booktitle    = {{EMNLP}},
  pages        = {25961--25970},
  publisher    = {Association for Computational Linguistics},
  year         = {2025}
}

@article{mem1,
  author       = {Zijian Zhou and
                  Ao Qu and
                  Zhaoxuan Wu and
                  Sunghwan Kim and
                  Alok Prakash and
                  Daniela Rus and
                  Jinhua Zhao and
                  Bryan Kian Hsiang Low and
                  Paul Pu Liang},
  title        = {{MEM1:} Learning to Synergize Memory and Reasoning for Efficient Long-Horizon
                  Agents},
  journal      = {CoRR},
  volume       = {abs/2506.15841},
  year         = {2025}
}

@article{memagent,
  author       = {Hongli Yu and
                  Tinghong Chen and
                  Jiangtao Feng and
                  Jiangjie Chen and
                  Weinan Dai and
                  Qiying Yu and
                  Ya{-}Qin Zhang and
                  Wei{-}Ying Ma and
                  Jingjing Liu and
                  Mingxuan Wang and
                  Hao Zhou},
  title        = {MemAgent: Reshaping Long-Context {LLM} with Multi-Conv RL-based Memory
                  Agent},
  journal      = {CoRR},
  volume       = {abs/2507.02259},
  year         = {2025}
}

@article{cognee,
  author       = {Vasilije Markovic and
                  Lazar Obradovic and
                  L{\'{a}}szl{\'{o}} Hajdu and
                  Jovan Pavlovic},
  title        = {Optimizing the Interface Between Knowledge Graphs and LLMs for Complex
                  Reasoning},
  journal      = {CoRR},
  volume       = {abs/2505.24478},
  year         = {2025}
}

@online{claudecode,
  author    = {},
  title     = {(Anthropic)},
  year      = {Claude Code},
  url       = {https://www.claude.com/product/claude-code}
}


\end{document}